\documentclass[lettersize,journal]{IEEEtran}
\usepackage{amsmath,amsfonts}
\usepackage{algorithmic}
\usepackage{algorithm}
\usepackage{array}
\usepackage[caption=false,font=normalsize,labelfont=sf,textfont=sf]{subfig}
\usepackage{textcomp}
\usepackage{stfloats}
\usepackage{url}
\usepackage{verbatim}
\usepackage{graphicx}
\usepackage{cite}
\usepackage{multirow}
\begin{document}

\title{Trajectory Planning for Autonomous Driving in
Unstructured Scenarios Based on Graph Neural Network and Numerical Optimization}

\author{Sumin Zhang, Kuo Li, Rui He, Zhiwei Meng, Yupeng Chang, Xiaosong Jin, Ri Bai
\thanks{The authors are with the School of National Key Laboratory of Automotive Chassis Integration and Bionics, Jilin University, Changchun 130012, China (e-mail: suminzhang@163.com; likuo23@mails.jlu.edu.cn; herui@jlu.edu.cn; mengzw20@mails.jlu.edu.cn; changyp21@mails.jlu.edu.cn; jinxs21@mails.jlu.edu.cn; bairi22@mails.jlu.edu.cn).}
}

\markboth{}%
{Shell \MakeLowercase{\textit{et al.}}: A Sample Article Using IEEEtran.cls for IEEE Journals}


\maketitle

\begin{abstract}
In unstructured environments, obstacles are diverse and lack lane markings, making trajectory planning for intelligent vehicles a challenging task. Traditional trajectory planning methods typically involve multiple stages, including path planning, speed planning, and trajectory optimization. These methods require the manual design of numerous parameters for each stage, resulting in significant workload and computational burden. While end-to-end trajectory planning methods are simple and efficient, they often fail to ensure that the trajectory meets vehicle dynamics and obstacle avoidance constraints in unstructured scenarios. Therefore, this paper proposes a novel trajectory planning method based on Graph Neural Networks (GNN) and numerical optimization. The proposed method consists of two stages: (1) initial trajectory prediction using the GNN, (2) trajectory optimization using numerical optimization. First, the graph neural network processes the environment information and predicts a rough trajectory, replacing traditional path and speed planning. This predicted trajectory serves as the initial solution for the numerical optimization stage, which optimizes the trajectory to ensure compliance with vehicle dynamics and obstacle avoidance constraints. We conducted simulation experiments to validate the feasibility of the proposed algorithm and compared it with other mainstream planning algorithms. The results demonstrate that the proposed method simplifies the trajectory planning process and significantly improves planning efficiency.
\end{abstract}

\begin{IEEEkeywords}
Autonomous driving, graph neural network(GNN), trajectory planning, numerical optimal control.
\end{IEEEkeywords}

\section{Introduction}
\IEEEPARstart{T}{rajectory} planning for intelligent vehicles is a crucial topic in the field of autonomous driving, requiring consideration of safety, comfort, and driving efficiency\cite{10329446}. Current trajectory planning research mainly focuses on structured environments such as highways and unstructured environments such as parking lots. In contrast to structured environments, which have lane markings as driving references, unstructured environments generally lack such references and feature more complex and diverse obstacles, making trajectory planning significantly more challenging\cite{9531561,10104141}. A feasible trajectory should connect the vehicle's initial position and orientation with its target position and orientation, comply with vehicle dynamic constraints, and ensure that the control system avoids collisions with obstacles during trajectory tracking. 

The objective of the trajectory planner is to establish a correspondence between the unstructured environment and the trajectory, generating a feasible path based on the environmental information. Traditional trajectory planning algorithms employ mathematical methods to identify such correspondences, typically involving multiple stages of mathematical modeling\cite{10104141,9531561}. These stages include path planning, path optimization, and speed planning to derive a feasible trajectory. This paper aims to simplify this process by leveraging deep learning techniques to identify the relationship between environmental information and the trajectory through large amounts of data. Subsequently, an optimal control method is used to generate a locally optimal trajectory that satisfies multiple constraints.

\section{Related Works}
Existing trajectory planning algorithms can be categorized based on different criteria. In line with the research content of this paper, trajectory planning methods are divided into three categories: 1) combinations of traditional planning algorithms, 2) optimization-based methods, and 3) end-to-end methods.

Traditional planning algorithms, originating in the 1970s, have been widely applied in the field of robotics\cite{10363676}. These algorithms typically seek a feasible or optimal solution based on certain manually defined rules. For example, graph search-based algorithms such as Dijkstra's\cite{dijkstra2022note} and A*\cite{hart1968formal}, and sampling-based algorithms such as RRT*\cite{lavalle2001randomized} and its variants, are commonly used. The paths solved by these algorithms often exhibit discontinuous curvature and do not consider vehicle kinematic characteristics, making them more suitable for global path planning. When used for local path planning, polynomial curves like quintic polynomials or Bézier curves are often used to fit the planned path. To simulate vehicle kinematics, the Hybrid A*\cite{dolgov2008practical} extends graph search algorithms by incorporating the vehicle's movement direction, resulting in smoother paths that adhere to vehicle kinematic constraints. To generate a trajectory, it is necessary to add time information to the path to represent the vehicle's speed. Speed planning typically employs S-T graph-based algorithms. The combination of traditional planning algorithms usually requires the manual setting of multiple stages. For instance, in graph search algorithms, the map must be gridded, and the grid resolution must be set. Larger grids can lead to poorer path quality, while smaller grids increase the computational cost of the algorithm. In sampling-based algorithms, the step size and the number of iterations need to be set. A smaller step size and more iterations can improve path quality but also increase computational cost, and vice versa.

Optimization-based methods generally refer to the design of a numerical optimization cost function with objectives such as comfort or energy consumption, and constraints such as collision avoidance or physical limitations. By minimizing this cost function, a locally optimal trajectory can be obtained. For example, Lim et al.\cite{8242694} used a hierarchical trajectory planning method combining sampling and numerical optimization. They obtained a rough behavioral trajectory through a sampling algorithm, then established a cost function and constraints, and performed numerical optimization using Sequential Quadratic Programming (SQP)\cite{boggs1995sequential}. When the continuous variables of the optimization problem are partially or fully discretized and then solved directly, it is typically referred to as an optimal control method\cite{1018171435.nh}. Optimal control methods model the trajectory planning problem as an Optimal Control Problem (OCP), usually consisting of an objective function and multiple constraints. The variables are discretized and converted into a Nonlinear Programming (NLP) problem for solving, using methods such as SQP and Interior Point Methods (IPM)\cite{wachter2006implementation}. Li et al.\cite{li2015unified} and Li and Wang\cite{7463491}, for example, formulated the trajectory planning task with constraints on vehicle dynamics, obstacles, and initial positions, optimizing performance metrics such as trajectory smoothness and obstacle avoidance to obtain an optimal trajectory. Lian et al.\cite{10104141} proposed a two-stage trajectory planning method. In the first stage, they used an improved Hybrid A* algorithm to obtain an initial trajectory. In the second stage, they performed segmented optimization of the optimal control problem. This method achieved good results in narrow parking environments by reducing optimization time.

End-to-end autonomous driving technology can be distinguished in terms of learning methods as imitation learning (IL) and reinforcement learning (RL). IL learns from a large number of expert demonstrations, mimicking expert behavior across various driving scenarios, while RL accumulates rewards or penalties through interaction with the environment, aiming to maximize cumulative rewards\cite{10258330}. Although RL does not require pre-prepared training data, its efficiency during the training process tends to be lower. End-to-end autonomous driving traces its origins back to ALVINN\cite{pomerleau1988alvinn} in 1988, which utilized a fully connected neural network to process data from cameras and radar sensors and output steering values. In recent years, with the continuous advancement in the field of deep learning and the increase in computational power, various methods for end-to-end autonomous driving have emerged, exhibiting differences in input formats, intermediate processes, and output formats. Hawke et al.\cite{9197408} trained a neural network model using real-world traffic environment data, inputting data from three different directions of cameras and directly outputting control signals. TransFuser\cite{9578103,9863660} utilized the attention mechanism from Transformers\cite{vaswani2017attention}, combining camera inputs and LIDAR data through multimodal fusion within the CARLA simulator\cite{dosovitskiy2017carla}, outputting 2D waypoints, and utilizing PID controllers for lateral and longitudinal control. PlanT\cite{renz2022plant} takes target-level environment as input, employs Transformers for fusion, and outputs 2D waypoints. It visualizes attention weights to demonstrate the interpretability of driving decisions. Some studies modularize end-to-end models, such as Sadat et al.\cite{sadat2020perceive} and Hu et al.\cite{hu2023planning}, who predict the motion trends of other participants in the environment between the model input and output, enhancing safety during model planning and interpretability during decision-making. IL requires large datasets during training to enhance the model's generalization ability but may struggle to adapt to significantly different new scenarios from the training dataset. To alleviate the limitations posed by insufficiently comprehensive datasets, some studies enrich training datasets by acquiring driving data from other vehicles on the road. For instance, Zhang et al.\cite{zhang2021learning} and Chen et al.\cite{chen2022learning} create more diverse driving datasets by capturing the driving behaviors of other vehicles.

There are also studies that combine the aforementioned methods. For instance, Wang et al.\cite{9037111} proposed the Neural RRT* algorithm, which utilizes neural networks to predict the probability distribution of the optimal path on the map and employs RRT* for searching. Zhang et al.\cite{9582813} enhanced the performance of Neural RRT* in predicting the probability distribution of the optimal path by utilizing Generative Adversarial Networks (GAN). Although such methods enhance path planning efficiency, the planned paths cannot be directly followed by the vehicle control. Zhao et al.\cite{9756640} employed GAN to learn the relationship between starting points, endpoints, and sequences of control actions in unstructured road environments. They integrated the RRT algorithm to extend the planned trajectory length. Although the algorithm can generate high-quality trajectories, they are not necessarily optimal. 
Li et al.\cite{10310161} utilized convolutional neural networks to predict priority search areas in grid maps, followed by Monte-Carlo tree search (MCTS) for path planning. After path optimization and velocity planning, a feasible trajectory is generated. Du et al.\cite{du2024lane} addressed lane-changing scenarios by using two neural networks: one predicts the feasibility of lane-changing, while the other predicts the lane-changing time and longitudinal displacement to determine the lane-changing trajectory

Based on the above methods, in unstructured environments, traditional trajectory planning methods, although reliable, necessitate manual design of algorithm parameters and often result in high time and space complexity during the solving process. Trajectory planning methods based on optimal control can generate high-quality trajectories while considering vehicle kinematic constraints, however, they rely on a feasible initial solution to reduce optimization time. The quality of the initial feasible solution directly impacts the optimization rate\cite{10104141,li2015unified,7463491}. In unstructured environments, end-to-end trajectory planning requires longer trajectories compared to structured environments. The end-to-end prediction trajectory is difficult to satisfy the vehicle's kinematics and has error fluctuations, making it unsuitable for vehicle control module tracking. Additionally, obstacles in such environments exhibit diverse forms, necessitating a fusion of complementary approaches with traditional methods to accomplish the task\cite{9037111,9582813,9756640}.

Drawing upon the advantages and disadvantages of the aforementioned algorithms, this study proposes a novel trajectory planning approach that integrates the concise and efficient nature of end-to-end methods with the ability of optimal control methods to learn driving trajectories while considering vehicle kinematic constraints and collision avoidance constraint. We employ a neural network model to not only learn 2D path point information but also the control inputs and state information of the vehicle, including vehicle position, heading angle, velocity, and front-wheel steering angle. The output values of the neural network serve as the initial solution for the optimal control algorithm, which refines them to yield a locally optimal trajectory. Simultaneously, the locally optimal solution serves as the ground truth for training the neural network. From the perspective of optimal control, the neural network efficiently provides an initial solution, thereby accelerating the optimization iterations of the optimal control problem. From the standpoint of end-to-end trajectory planning, the optimal control proposition ensures that the trajectory adheres to the vehicle's kinematic constraints, facilitating the tracking by the control system. Consequently, this research presents a novel approach to trajectory planning by combining end-to-end trajectory planning with optimal control propositions, leveraging their complementary strengths to enhance planning efficiency and trajectory feasibility.
\begin{figure*}[htb]
    \centering
    \includegraphics[width=0.8\textwidth]{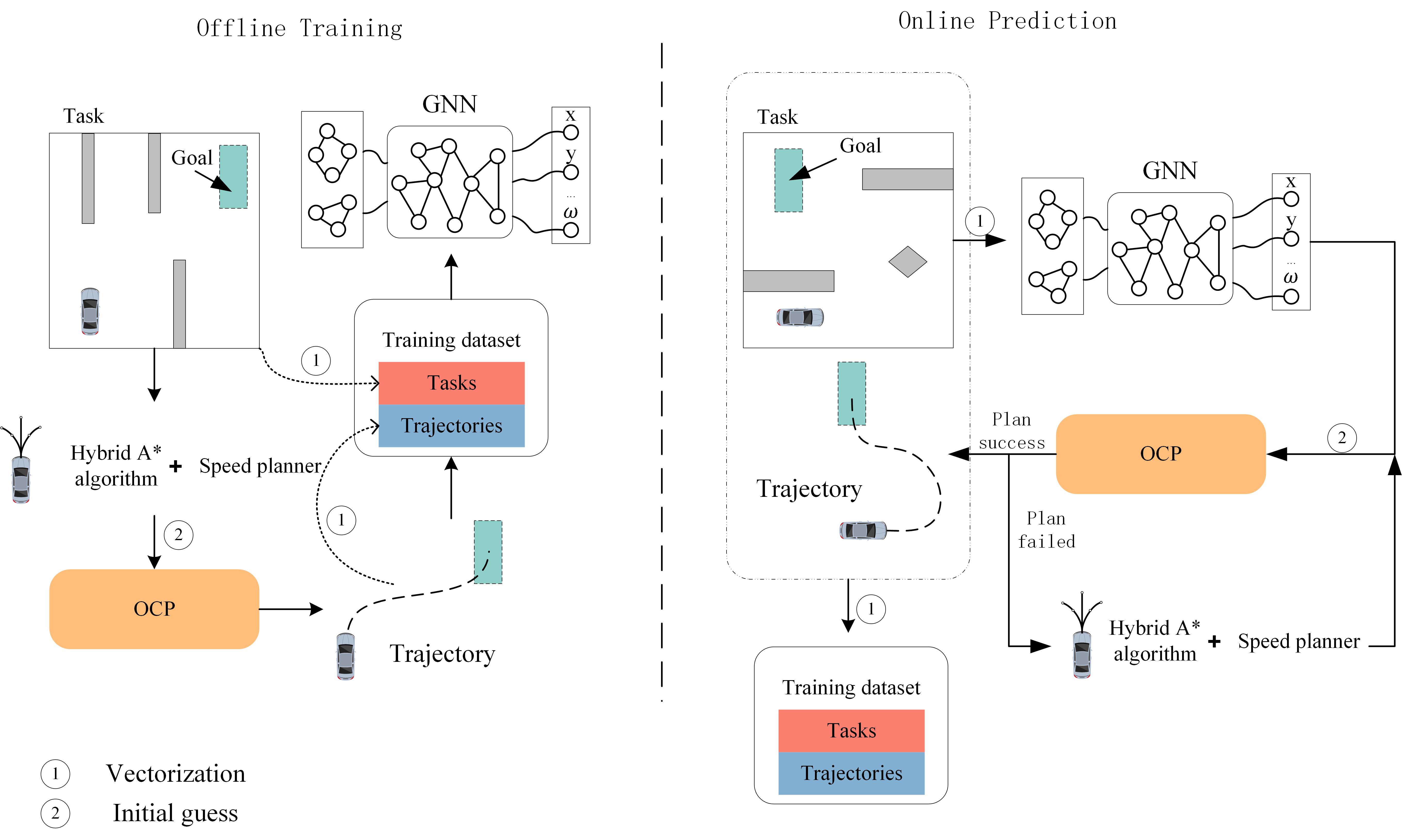}
    \caption{Overall framework}
    \label{fig:Overall framework}
\end{figure*}

\subsection{Contributions}

The main contributions of this paper are as follows:

1) We integrate neural networks with optimal control to devise a novel trajectory planning architecture, demonstrating the feasibility of this approach and enhancing both the efficiency and quality of trajectory planning.

2) We utilize graph neural networks to extract and encode information from unstructured environments, leading to a more streamlined and efficient neural network structure.

3) Neural networks predict trajectories as initial solutions for optimal control, simplifying the process of obtaining initial solutions and thereby improving trajectory planning efficiency.

4) Optimal control optimizes the trajectory predicted by the neural network, ensuring compliance with vehicle kinematic constraints and obstacle avoidance constraints.

\subsection{Organization}
The remaining structure of this paper is outlined as follows.
In Section \ref{System Framework Overview}, we provide a brief description of trajectory planning propositions and introduce the algorithmic framework of this study. Section \ref{Proposed Trajectory Planner} elaborates on the neural network structure and optimal control algorithm employed in the planning algorithm. Section \ref{Experiment Analysis} presents the implementation details of our algorithm and compares it with mainstream planning algorithms through experimental evaluation. Section \ref{Conclusion} concludes our research findings.

\section{System Framework Overview}
\label{System Framework Overview}
\subsection{Trajectory Planning Problem}
The core objective of trajectory planning lies in identifying a safe and feasible trajectory given known environmental information and the starting point. Vehicle trajectory planning must initially satisfy obstacle avoidance constraints and vehicle kinematic constraints.
In this paper, a trajectory is defined as $\sigma$. The trajectory should connect the initial state $x_{init}$ with the goal state $x_{final}$. Under a certain cost function, the trajectory should be locally optimal. Therefore, the path planning proposition can be defined as follows:
\begin{equation}
\begin{aligned}
&\sigma^*=\underset {\sigma \in \sum} { \operatorname {arg\,min} } \,c(\sigma)\\
s.t. \quad &\sigma(0)=x_{init}\\
&\sigma(t_{final})=x_{final}\\
&\sigma(t)\in \mathcal{X}_{free}\quad\quad \forall  t \in [0, t_{final}]
\end{aligned}
\end{equation}
$\sum$ represents the set of all feasible trajectories, $\sigma^*$ denotes the optimal trajectory, and $\mathcal{X}_{free}$ signifies the free space where collision with obstacles is avoided.
\begin{figure*}[htb]
    \centering
    \includegraphics[width=0.8\textwidth]{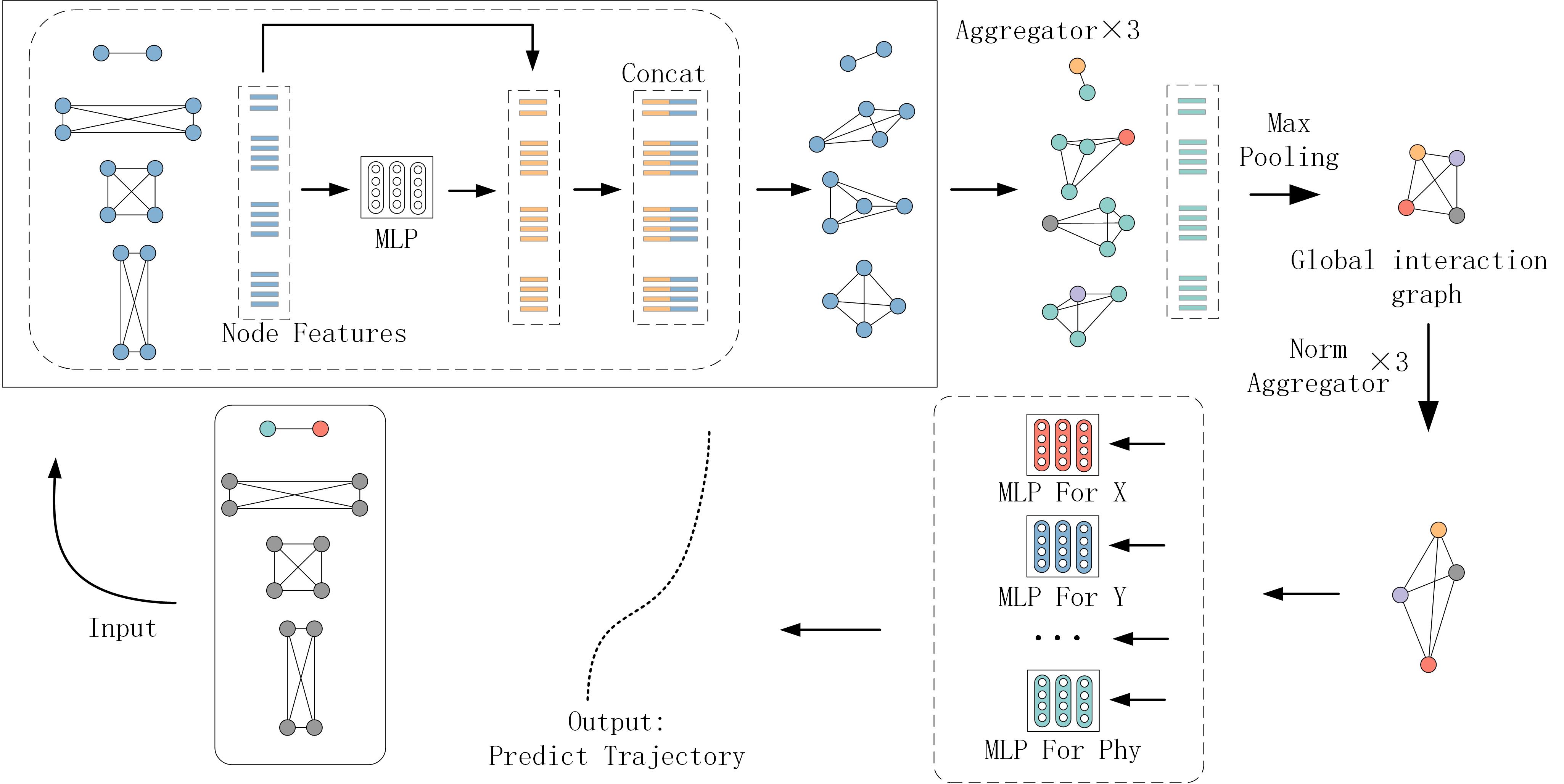}
    \caption{Network architecture schematic diagram}
    \label{fig:net}
\end{figure*}
\subsection{System Framework}
Based on graph neural networks and the optimal control trajectory planning algorithm, as depicted in Fig. \ref{fig:Overall framework}, the process consists of two operational phases: offline training and online prediction.
\subsubsection*{offline training}
Offline training aims to enable the neural network to learn how to generate an initial trajectory under various tasks, serving as the initial solution for the optimal control problem. We prepared a large number of environmental maps and randomly specified the initial and target positions of the vehicle. Each pair of starting positions and environmental configurations constitutes a task. Subsequently, we utilized the Hybrid A* algorithm to solve each task and obtained feasible trajectories through velocity planning. These feasible trajectories serve as the initial solutions for the optimal control problem, which are refined to yield locally optimal trajectories through optimization. The combination of tasks and trajectories forms a sample, and the accumulation of all samples constitutes the training dataset for the neural network.
\subsubsection*{online prediction}
Online prediction involves replacing the path planning and velocity planning stages with the trained neural network to enhance planning efficiency. The starting position, target position, and environment configuration of vectorization are used as inputs for graph neural networks. Through network inference, the predicted vehicle trajectory is outputted. This predicted trajectory serves as the initial solution for the optimal control problem, which is further refined to yield a locally optimal trajectory through optimization. The output of the neural network prediction determines the success of the optimal control problem's solution and the efficiency of finding the optimal value. High-quality prediction results will shorten the solution time of the optimal control problem. If the optimal control problem fails to converge, a trajectory will be solved using the rule-based Hybrid A* and Speed planner before reattempting optimization.

\section{Proposed Trajectory Planner}
\label{Proposed Trajectory Planner}
In this section, we introduce two critical phases of trajectory planning in this paper: 1) employing neural networks to predict trajectories as initial solutions for OCP, and 2) optimizing the initial solutions provided by the neural networks to generate a locally optimal trajectory. The specific details are outlined below.
\subsection{Learning-based Initial Solution Generation}
Our neural network architecture draws inspiration from VectorNet\cite{gao2020vectornet} and PlanT. Unlike CNN models that take grid-based pixel points as input, we represent tasks using object-level representations and extract features through Graph Neural Networks (GNNs). In addition, our network structure is relatively simple and there is enough room for improvement. The network structure is depicted in Fig. \ref{fig:net}.
\subsubsection*{Task vectorization and subgraph generation}
\begin{figure}[htb]
    \centering
    \includegraphics[width=0.45\textwidth]{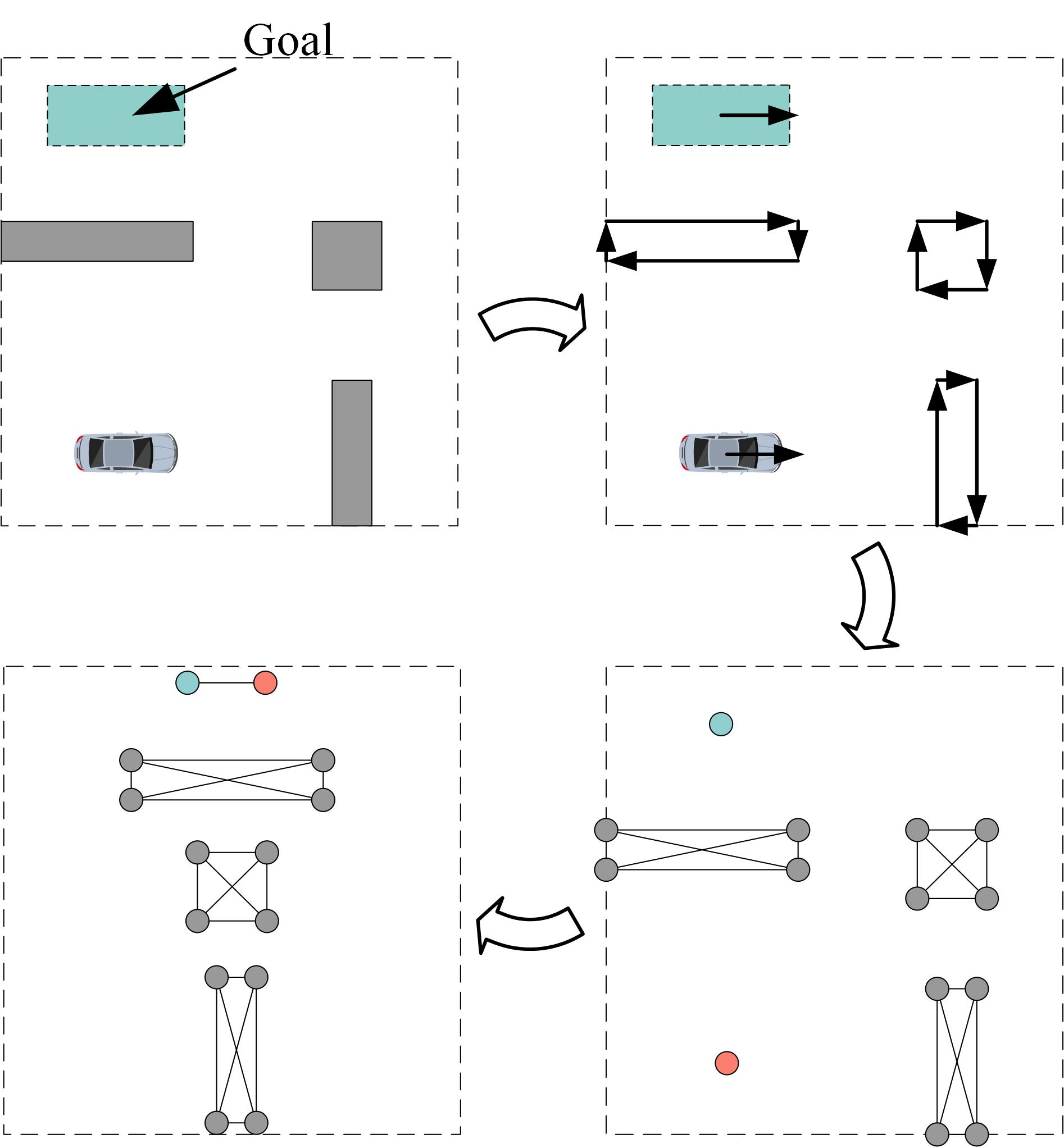}
    \caption{Vectorization}
    \label{fig:Vectorization}
\end{figure}

Before vectorization, to standardize the initial state of each task, we convert the absolute positions of obstacles and target states in the Cartesian coordinate system into relative positions with respect to the initial state. The vectorization process is illustrated in Fig. \ref{fig:Vectorization}. The start and end vectors contain the vehicle's initial coordinates and heading angle, forming the starting vector $v^s$ and the target vector $v^f$. The vertices of each obstacle are connected to each other to form obstacle vectors $v_i^o$. To enable the GNN to efficiently acquire task information, we interconnect obstacle vectors, with each obstacle forming an obstacle subgraph. Simultaneously, the starting point and the endpoint form a state subgraph. All subgraphs constitute the task graph.

\subsubsection*{Neural network structure}
To fully utilize the positional information contained within each subgraph, we perform a three-layer feature fusion for each subgraph, which includes the feedforward phase of a fully connected neural network and graph feature concatenation. The single-layer feature fusion propagated in each subgraph can be defined as follows:
\begin{equation}
\label{eq1}
G_{i}^{(l+1)} = {\gamma}_{con}({\varphi}_{mlp}(G_{i}^{(l)}),{\varphi}_{agg}(\{{\varphi}_{mlp}(G_{i}^{(l)})\}))
\end{equation}
Here, $G_{i}^{(l+1)}$ represents the subgraph features of the $i$-th subgraph in the $l$-th layer. The ${\varphi}_{mlp}(·)$ is a multi-layer perceptron consisting of three fully connected layers with 128 neurons in the hidden layer. The ${\varphi}_{agg}(·)$ denotes the feature aggregation layer, and ${\gamma}_{con}(·)$ represents the feature concatenation layer. After three layers of feature fusion, to extract important features from the subgraphs within the task, we perform max pooling on each subgraph as follows:
\begin{equation}
\label{eq2}
G_{i}^{(l+1)}={\varphi}_{maxpool}(G_{i}^{(l)})
\end{equation}
$\varphi_{maxpool}(·)$ represents the max pooling layer.

The max-pooled vector for each subgraph represents its most salient features. These features are concatenated to form a new graph structure, which then undergoes three rounds of feature aggregation to complete the feature extraction of the task graph.

Following the feature fusion and extraction of the task graph, to reduce the size of the network, the structure only predicts five variables that constitute the vehicle's trajectory, denoted as $T=\{x_j,y_j,\theta_j,v_j,{\varphi}_j\}$. Here, $x_j,y_j,{\theta}_j,v_j,{\varphi}_j$ represent the position coordinates, heading angle, speed, and front-wheel steering angle at the $j$-th trajectory point, respectively. Unpredicted variables, such as vehicle acceleration $a$ and front-wheel steering angular velocity $w$, are computed from $v$ and $\varphi$.

\subsection{Optimize the Initial Trajectory}
In this section, based on previous research \cite{9531561,9729517,7463491}, we formulate an OCP using neural network outputs as the initial solution. This includes defining the cost function to be minimized and the various constraints to be applied. Finally, we briefly introduce the solution approach for the OCP.
\subsubsection{Problem Overview}
Usually, the OCP for trajectory planning can be described in the following standard form:
\begin{equation}
\label{cost_fun}
min J(u(t),x(t),t_{final})
\end{equation}
\begin{equation}
\label{eq_all_constrain}
\begin{cases}
\dot{x}(t)=f_{kin}(u(t),x(t)) \\ 
x(0)=x_{init},x(t_{final})=x_{final} \\
u(0)=u_{init},u(t_{final})=u_{final} \\
u_{min}\leq u(t)\leq u_{max} \\
f_{path}(u(t),x(t)) \in \mathcal{X}_{free}
\end{cases}
\end{equation}
In this context, $J$ represents the cost function that includes both control and state variables. $\dot{x}(t)$ denotes the derivative of the state variable $x(t)$ with respect to time. $u(t)$ is the control variable that provides inputs to the vehicle. $f_{kin}()$ represents the relationship between the change in vehicle state and the current control variables, also known as the vehicle kinematic equation. $x_{init}$ and $x_{final}$ represent the initial and final states of the vehicle, respectively. The terms $u_{max}$ and $u_{min}$ denote the maximum and minimum control inputs. The workspace for trajectory planning is defined as $\mathcal{X}=\mathcal{X}_{free} \cup \mathcal{X}_{obs}$, where $\mathcal{X}_{obs}$ represents the obstacle space. The function $f_{path}()$ maps the vehicle's state and inputs to the workspace, presenting them as a path that should not intersect with obstacles, i.e., $f_{path}(u(t),x(t))\in \mathcal{X}_{free}=\mathcal{X}{\backslash}\mathcal{X}_{obs}$. The above defines the cost function and various constraints. Next, we will elaborate on these components in detail.
\begin{figure}[t]
    \centering
    \includegraphics[width=0.45\textwidth]{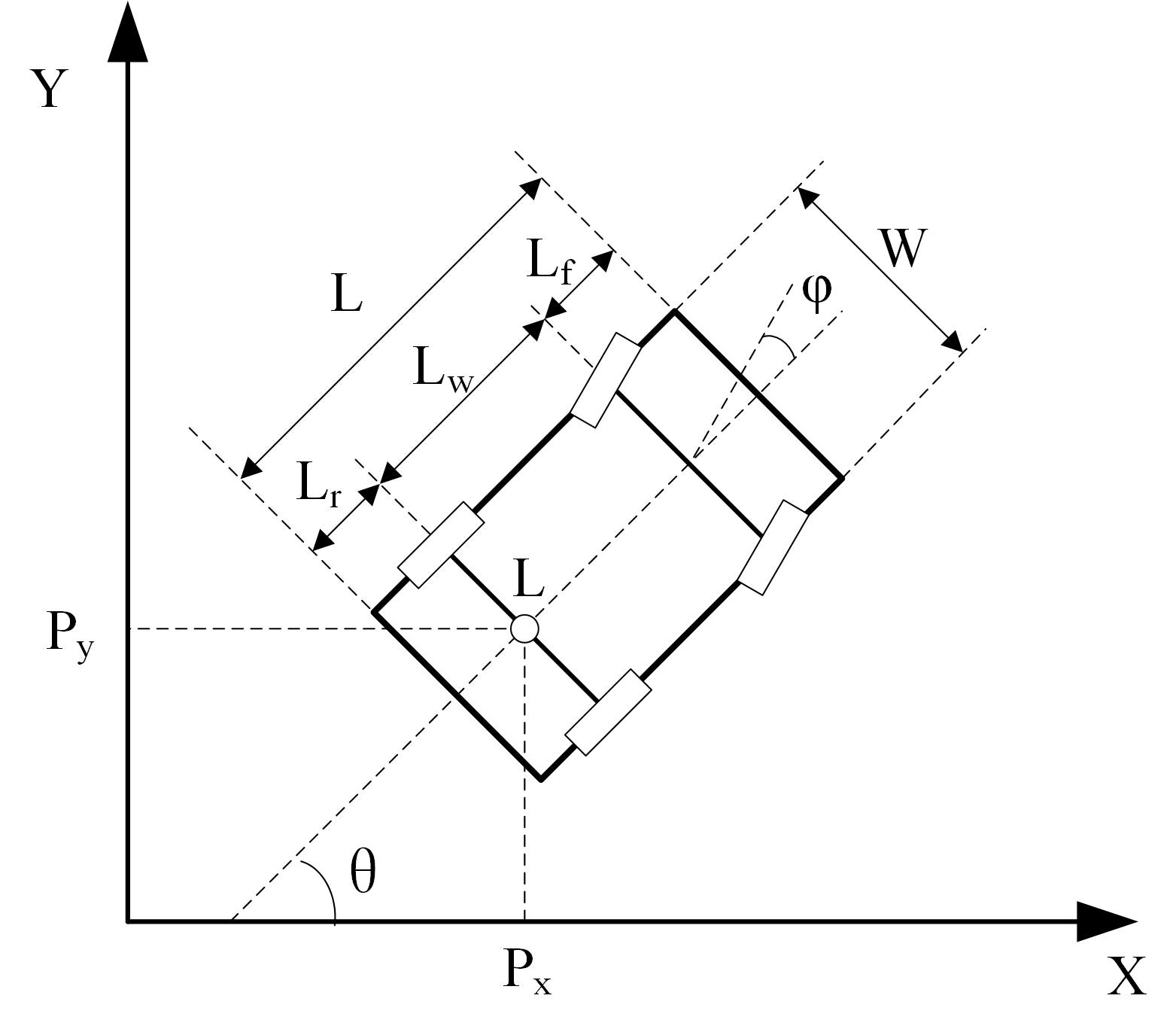}
    \caption{Geometric parameters related to vehicle kinematics.}
    \label{fig:Vehicle}
\end{figure}
\subsubsection{Various constraints} The constraints include kinematic constraints, initial and final state constraints, and obstacle constraints.
\subsubsection*{kinematic constraints}
In unstructured environments, vehicles typically need to operate at low speeds. Therefore, this paper employs the bicycle model to represent the vehicle's kinematics. A schematic diagram of the vehicle parameters is shown in Fig. \ref{fig:Vehicle}. The equation $\dot{x}(t)=f_{kin}(u(t),x(t))$ in equation \eqref{eq_all_constrain} is expanded as follows:
\begin{equation}
\begin{bmatrix} 
\dot{x}(t) \\ 
\dot{y}(t) \\
\dot{\theta}(t) \\
\dot{v}(t)\\
\dot{\varphi}(t)\\
\end{bmatrix} = 
\begin{bmatrix} 
{v}(t)·cos\theta(t) \\
{v}(t)·sin\theta(t) \\
{v}(t)·tan\ \varphi(t)/L_{w} \\
a_{t} \\
w_{t}
\end{bmatrix},t \in [0, t_{final}]
\end{equation}

$x(t)$ and $y(t)$ represent the coordinates of the vehicle's rear axle center in the Cartesian coordinate system. $\theta(t)$ denotes the vehicle's heading angle, while $v(t)$ and $a(t)$ represent the vehicle's velocity and acceleration, respectively. $\varphi(t)$ and $w(t)$ stand for the front-wheel steering angle and angular velocity. $L_w$ denotes the wheelbase of the vehicle. Additionally, in the practical operation of the vehicle, it should adhere to physical or mechanical constraints, satisfying:
\begin{equation}
\begin{bmatrix} 
-v_{max} \\ 
-a_{max} \\
-\varphi_{max} \\
-w_{max} \\
\end{bmatrix} \leq
\begin{bmatrix} 
{v}(t) \\
{a}(t) \\
\varphi(t) \\
w(t) \\
\end{bmatrix} \leq
\begin{bmatrix}
v_{max} \\ 
a_{max} \\
\varphi_{max} \\
w_{max} \\
\end{bmatrix}
\end{equation}
\subsubsection*{State constraints}
When solving the OCP problem, it's essential to meet the fundamental requirements of the vehicle's initial and final states in the trajectory planning task. As in equation \eqref{eq_all_constrain}, the initial and final states should satisfy:
\begin{equation}
x(0)=x_{init},x(t_{final})=x_{final}
\end{equation}
In this article, we assume that at the beginning and end of a trajectory planning task, the vehicle has no input effect, i.e
\begin{equation}
\begin{cases}
u(0)=u_{init}=0, \\
u(t_{final})=u_{final}=0
\end{cases}
\end{equation}
\subsubsection*{Obstacle constraints}
In unstructured environments, obstacles exhibit diverse forms, and the establishment of obstacle avoidance constraints is crucial for the completion of trajectory planning tasks. In previous research, \cite{li2015unified} utilized the triangle area method to establish collision avoidance constraints. To reduce the number of constraints, \cite{9531561} and \cite{10104141} introduced collision-free corridors, confining the vehicle within these corridors. In this paper, to simplify the processing procedure, we adopt the triangle area method to establish constraints.

\begin{figure}[htb]
    \centering
    \includegraphics[width=0.45\textwidth]{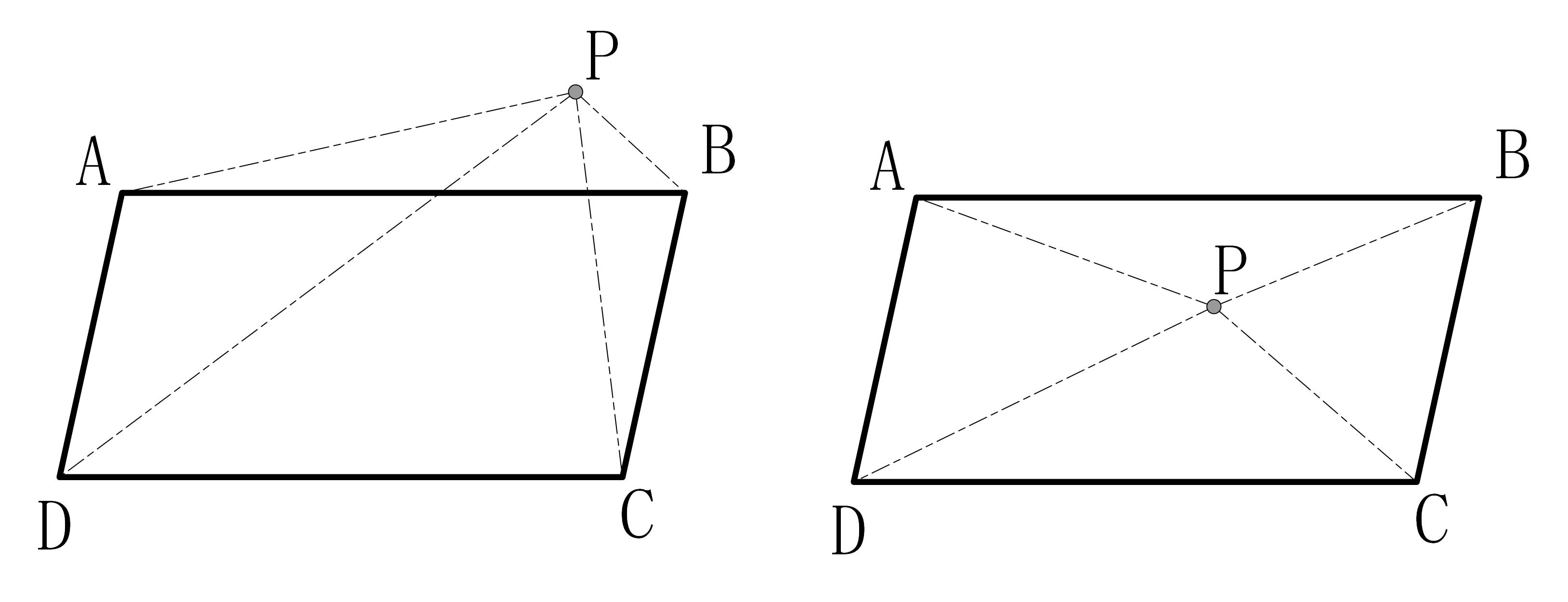}
    \caption{Triangle area method}
    \label{fig:Triangle area}
\end{figure}
The triangle area method is used to determine whether a point lies inside a polygon. As illustrated in Fig. \ref{fig:Triangle area}, the relationship holds for the triangle formed by point $P$ and the vertices of polygon $ABCD$:
\begin{equation}
\begin{matrix}
\sum S_{\triangle P_i}\geq S_{\square ABCD}, i=1 .. n \\ 
 \sum S_{\triangle Pi}=S_{\triangle PAB}+S_{\triangle PBC}+S_{\triangle PCD}+S_{\triangle PDA}
\end{matrix}
\end{equation}
When point $P$ lies inside polygon $ABCD$, then:
\begin{equation}
\sum S_{\triangle P_i}= S_{\square ABCD}, i=1 .. n \\ 
\end{equation}
otherwise:
\begin{equation}
\sum S_{\triangle P_i}> S_{\square ABCD}, i=1 .. n \\ 
\end{equation}
We establish collision avoidance constraints by considering the four vertices of the vehicle's quadrilateral and each obstacle polygon, as well as by considering each vertex of the obstacle polygons and the vehicle's quadrilateral. This dual approach ensures comprehensive collision avoidance.
\subsubsection{Cost function}
The purpose of establishing a cost function is to find a locally optimal solution that minimizes the cost within the set of constraint-satisfying solutions. Constructing the cost function for the trajectory planning problem should consider multiple criteria, including the smoothness of the vehicle's trajectory, safety, and the time required to complete the task. Different weight coefficients should be assigned to these criteria to reflect their relative importance.

To enhance the smoothness of the trajectory, sudden changes in speed and front-wheel steering angle should be minimized. This contributes to ride comfort and the effectiveness of the vehicle's control module in tracking the planned path. Therefore, let:
\begin{equation}
J_1 = \int_0^{t_{final}} w_{i}^2(t) +a_{i}^2(t)dt
\end{equation}
Regarding the safety of the trajectory, it should maintain a safe distance from obstacles. Assuming the geometric center of each obstacle is $C_j=(x_j^c,y_j^c)$, the performance metric can be designed as follows:
\begin{equation}
J_2 = \sum_{j=1}^{N_{obs}}(\int_0^{t_{final}}\gamma^{-d_{j}^2(t)}dt)
\end{equation}
Here, $d_{j}^2(t)=(x_i(t)-x_i^c)^2+(y_i(t)-y_i^c)^2$ represents the squared distance between the vehicle and the center of obstacle. The larger the distance, the smaller the cost function.
Assuming that each planned task consumes the same amount of time, i.e., $t_{final}$ is a constant, the $J$ in \eqref{cost_fun} can be expressed as:
\begin{equation}
J = \lambda_1·J_1 + \lambda_2·J_2
\end{equation}
$\lambda_1$ and $\lambda_2 $ represent the weight coefficients of the corresponding indicators.
\subsubsection{Solving OCP}
The solution of trajectory planning OCP requires the discretization of continuous variables, thus forming a NLP. When solving NLP, the establishment of initial solutions will affect whether the problem can be successfully solved and the speed of solving. Therefore, in previous studies, some scholars \cite{9531561,9729517,10104141} used the Hybrid A* algorithm and its variants as the basis to construct initial solutions. Building upon this foundation, this paper proposes the use of GNN to predict trajectories for constructing initial solutions. For NLP with existing initial solutions, we utilize the open-source solver IPOPT for solution, thereby completing the entire trajectory planning process. Some detailed basic parameters are listed in Table \ref{tab:BASIC_PARAMETERS}.

\begin{table}[!htbp]
\caption{BASIC PARAMETERS}
\label{tab:BASIC_PARAMETERS}
\begin{center}
\begin{tabular}{| 
>{\centering\arraybackslash}m{1.5cm}|
>{\centering\arraybackslash}m{4.5cm}|
>{\centering\arraybackslash}m{1.2cm}|}
 \hline
 Parameter&Description& Value\\ 
 \hline
    $L_w$ & Vehicle wheelbase& 2.8$m$\\ 
    \hline
    $L_f$  & Vehicle front hang length & 0.96$m$ \\ 
    \hline
    $L_f$ & Vehicle rear hang length& 0.929$m$\\ 
    \hline
    $W$ & Vehicle width& 1.9$m$\\ 
    \hline
    $a_{max}$ & Maximum acceleration& 2$m/s^2$\\ 
    \hline
    $w_{max}$ & Maximum angular velocity& 0.85$rad/s$\\ 
    \hline
    $v_{max}$ & Maximum velocity& 2.5$m/s$\\ 
    \hline
    $\varphi_{max}$ & Maximum steering angle& 0.7$rad/s$\\ 
    \hline
    $N_{out}$ & Number of GNN prediction points& 30\\  
    \hline
    $\Delta d_x$,$\Delta d_y$ & The xoy resolution for Hybrid A*& 0.2$m$,0.2$m$\\ 
    \hline
    $\Delta d_{theta}$ & Heading resolution for Hybrid A*& 0.785\\ 
    \hline
    $\lambda_1$ & Penalty for $J_1$& 1.0\\ 
    \hline
    $\lambda_1$ & Penalty for $J_2$& 1.0\\ 
    \hline
    $P_o$ & Number of optimization points & 120\\ 
\hline
\end{tabular}
\end{center}
\end{table}

\section{Experiment Analysis}
\label{Experiment Analysis}
\subsection{Implementation Details}
To validate the effectiveness of the proposed trajectory planning algorithm, a series of simulation experiments were conducted in typical scenarios. Detailed training procedures and simulation parameters are provided below.
\subsubsection{Training Details}
We prepared 800 maps with different random distributions of obstacles, and for each map, 15 pairs of random start and goal positions were selected. Each pair was solved using the Hybrid A* algorithm, and the resulting trajectory was optimized using IPOPT, forming one training sample. Due to the incompleteness of Hybrid A*, there were instances where no solution was found during task solving. After solving, a total of 10,426 local optimal trajectories were obtained as samples. 90\% of the samples were used as the training set, while the remaining samples were used as the test set. During training, the Adam optimizer was utilized with a learning rate of 0.001. Training was conducted on an RTX3090 GPU for 200 epochs, with a warm-up period of 50 epochs, and a batch size of 32.
\subsubsection{Simulation Details}
During algorithm simulation testing, to facilitate subsequent comparison with mainstream planning algorithms, GNN inference was conducted on an Intel i9-12900H CPU to align with other computations. The simulation platform utilized Python for implementation.
\subsection{Analysis of the Proposed Planner}
This section demonstrates the performance of the proposed method in different obstacle environments.

\begin{figure}[!t]
\centering
\subfloat[]
{\includegraphics[width=4.0cm]{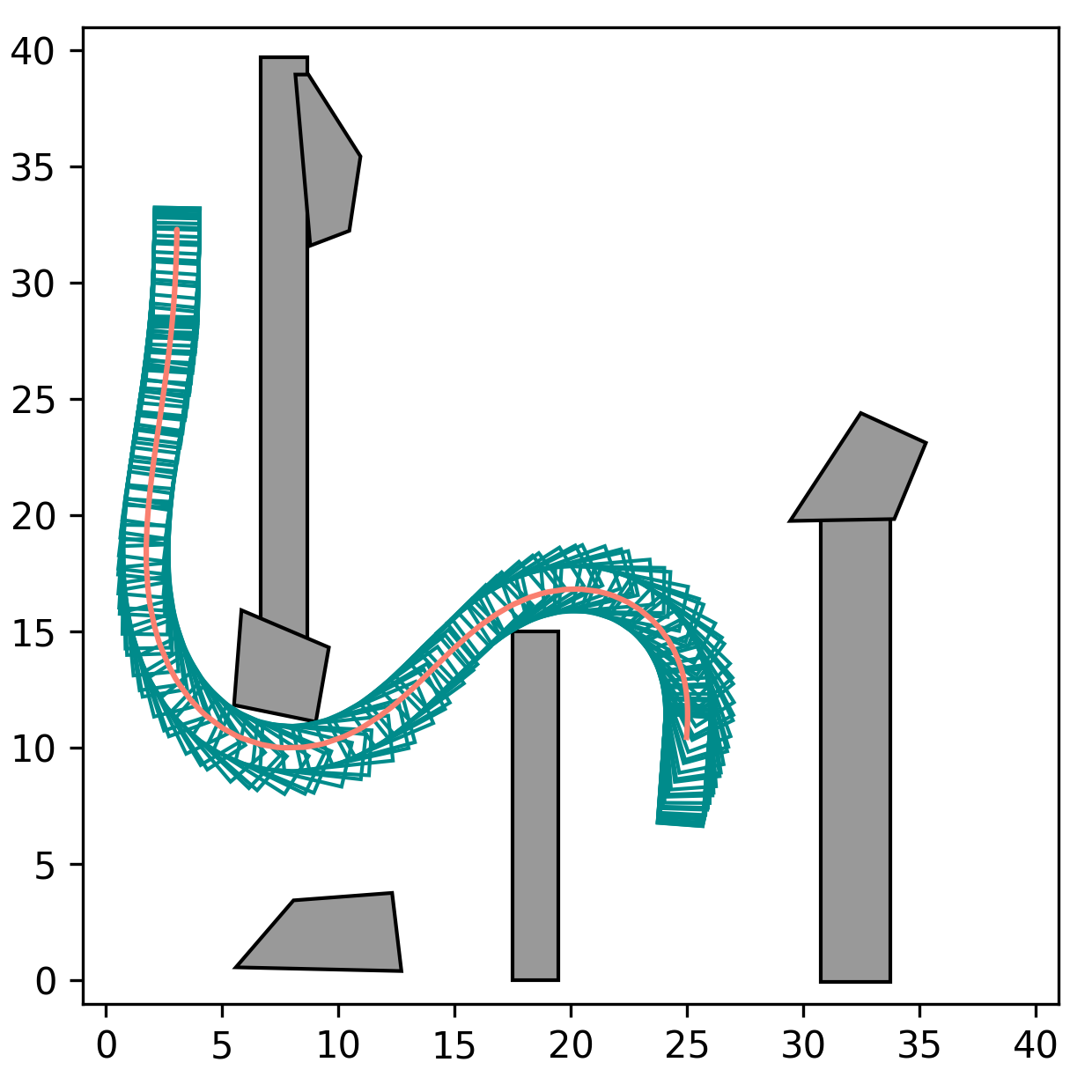}}
\hfil
\subfloat[]
{\includegraphics[width=4.0cm]{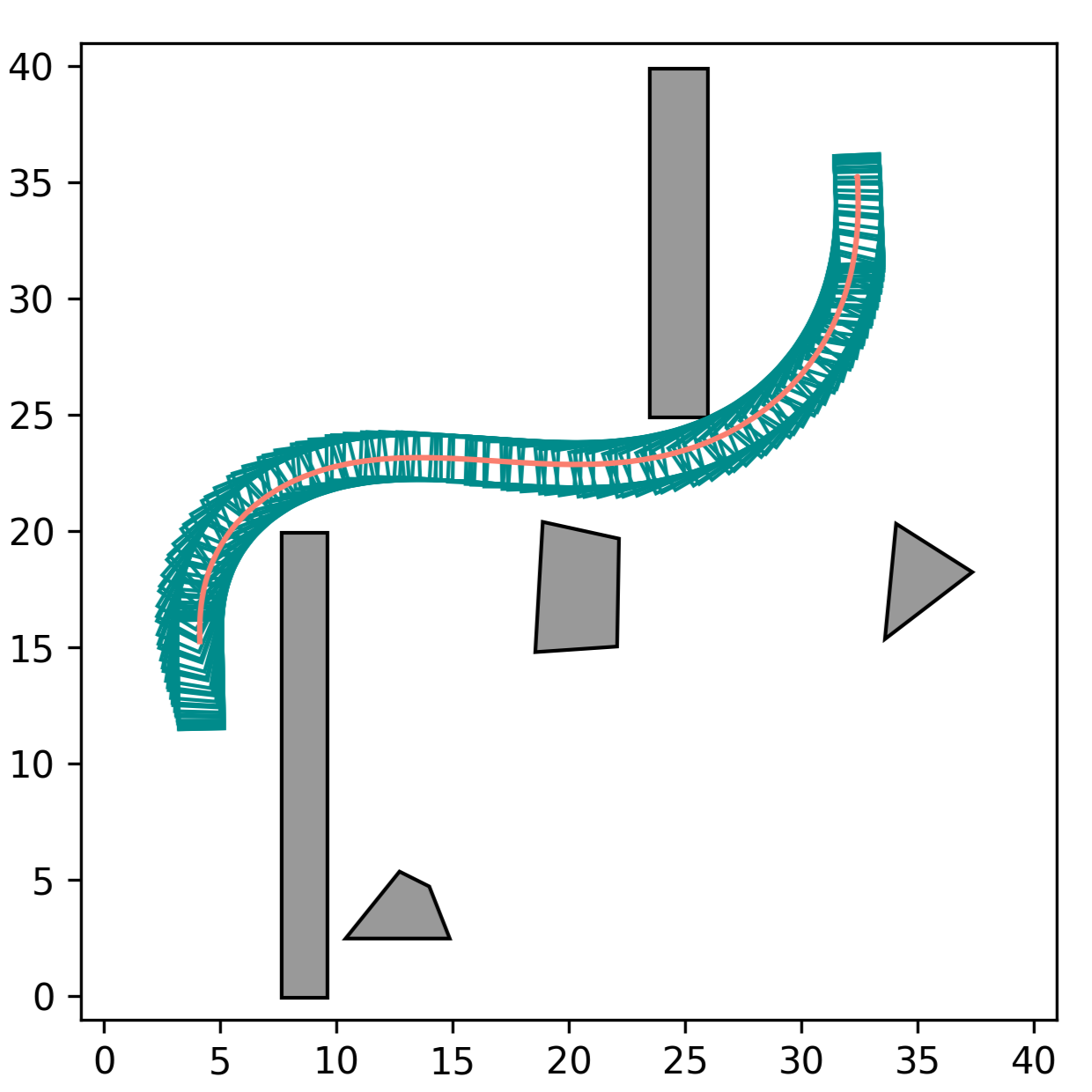}}
\hfil
\subfloat[]
{\includegraphics[width=4.0cm]{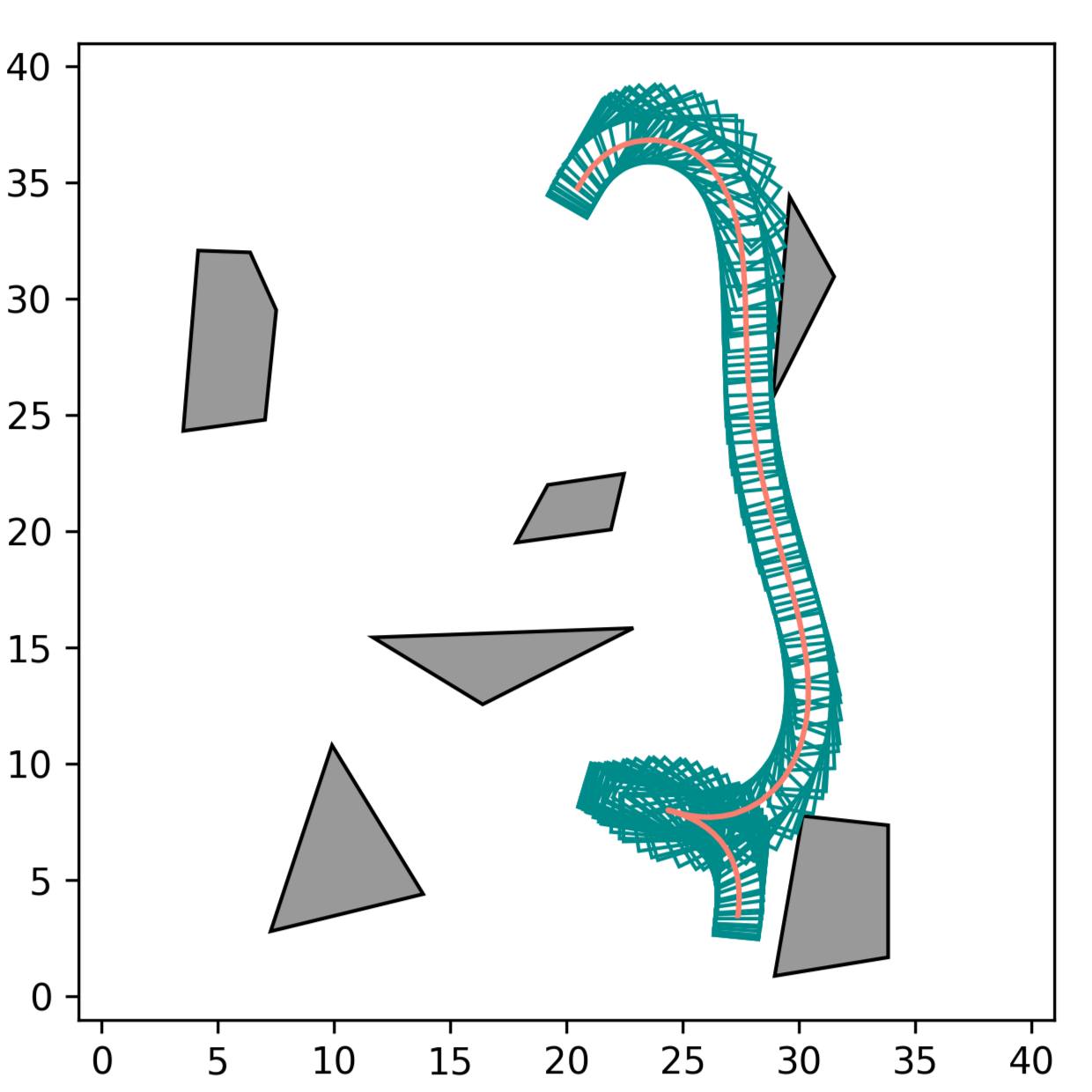}}
\hfil
\subfloat[]
{\includegraphics[width=4.0cm]{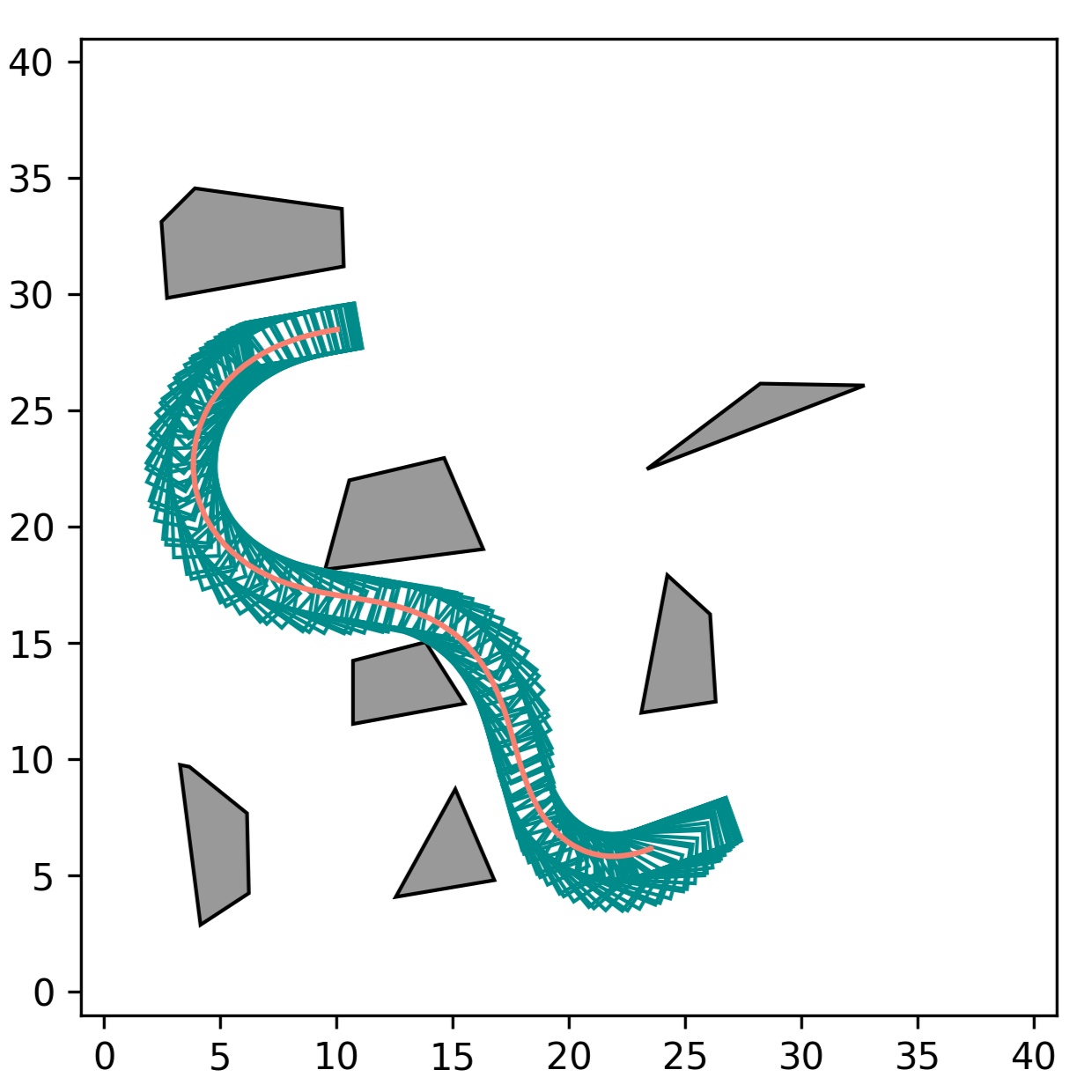}}
\hfil
\caption{Algorithm testing environment}
\label{fig:test_map}
\end{figure}

We tested the algorithm in four different obstacle environments as shown in Fig. \ref{fig:test_map}. \ref{fig:test_map}(a) and \ref{fig:test_map}(b) consist of rectangular obstacles and irregular obstacles, while \ref{fig:test_map}(c) and \ref{fig:test_map}(d) comprise several irregular obstacles. These scenarios highlight the algorithm's ability to plan longer trajectories and handle complex obstacles. In all four scenarios, the proposed algorithm successfully completes the trajectory planning tasks, with the green points in the figures representing the trajectory points formed by the vehicle.

\begin{figure}[htbp]
\centering
\subfloat[]
{\includegraphics[width=4.0cm]{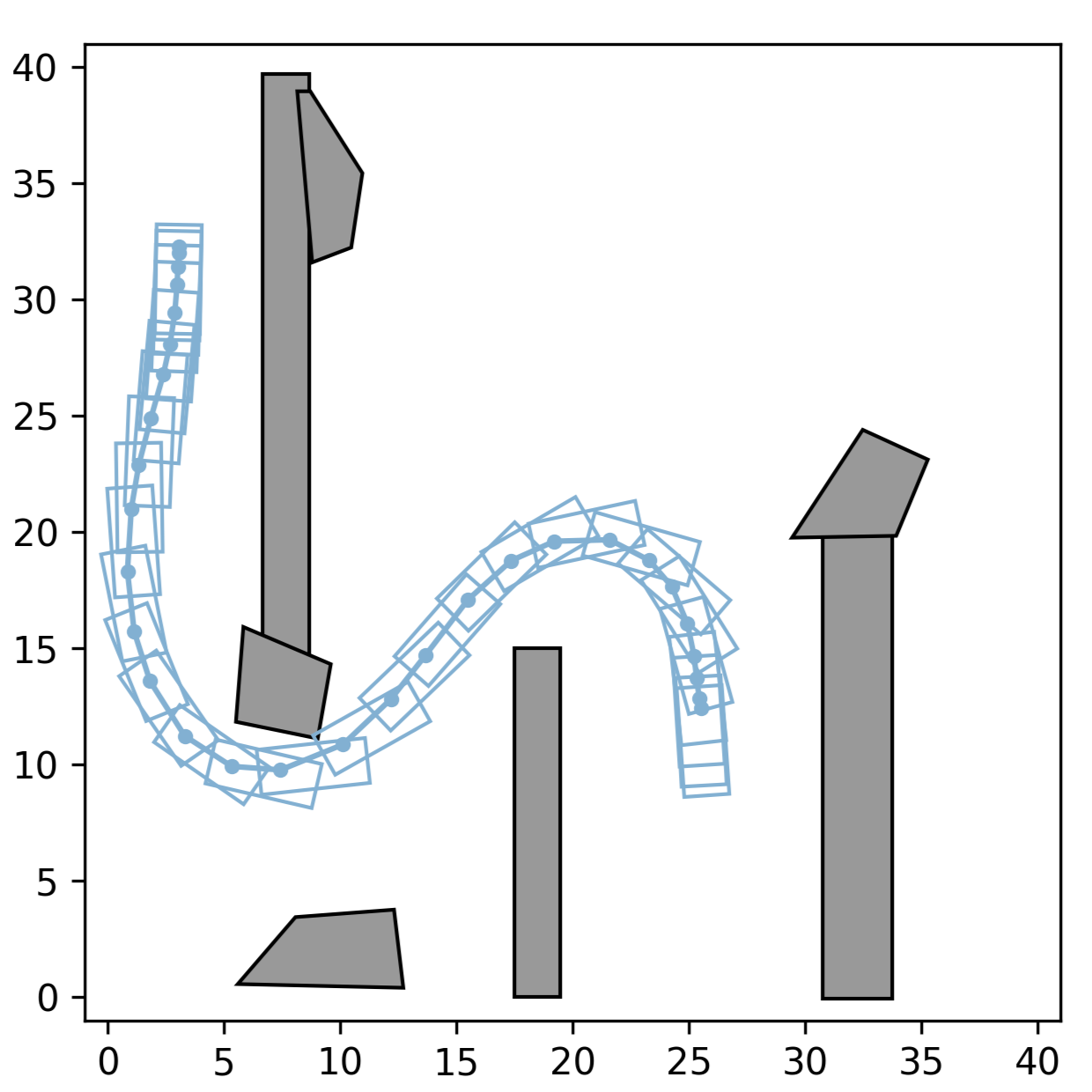}}
\hfil
\subfloat[]
{\includegraphics[width=4.0cm]{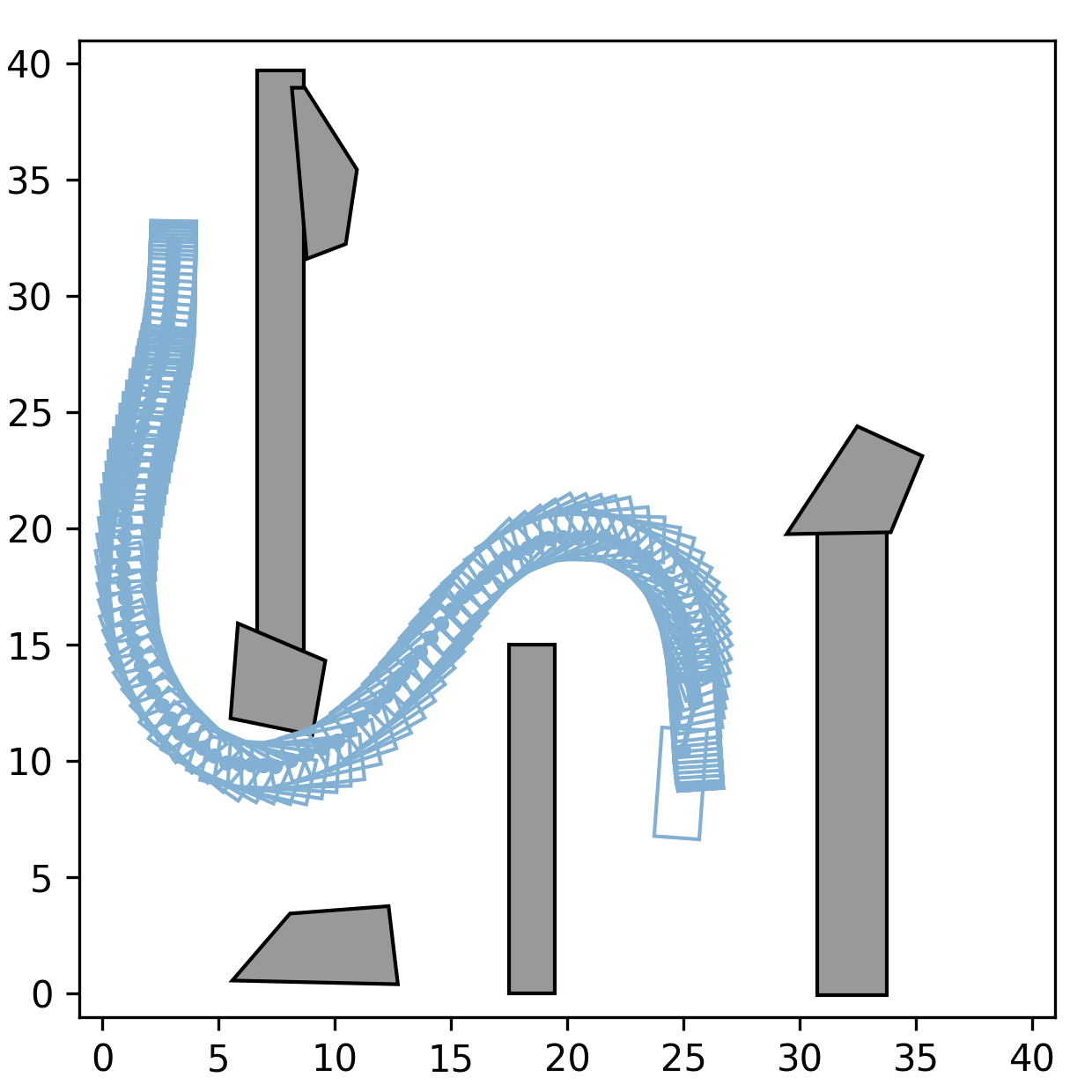}}
\hfil
\subfloat[]
{\includegraphics[width=4.0cm]{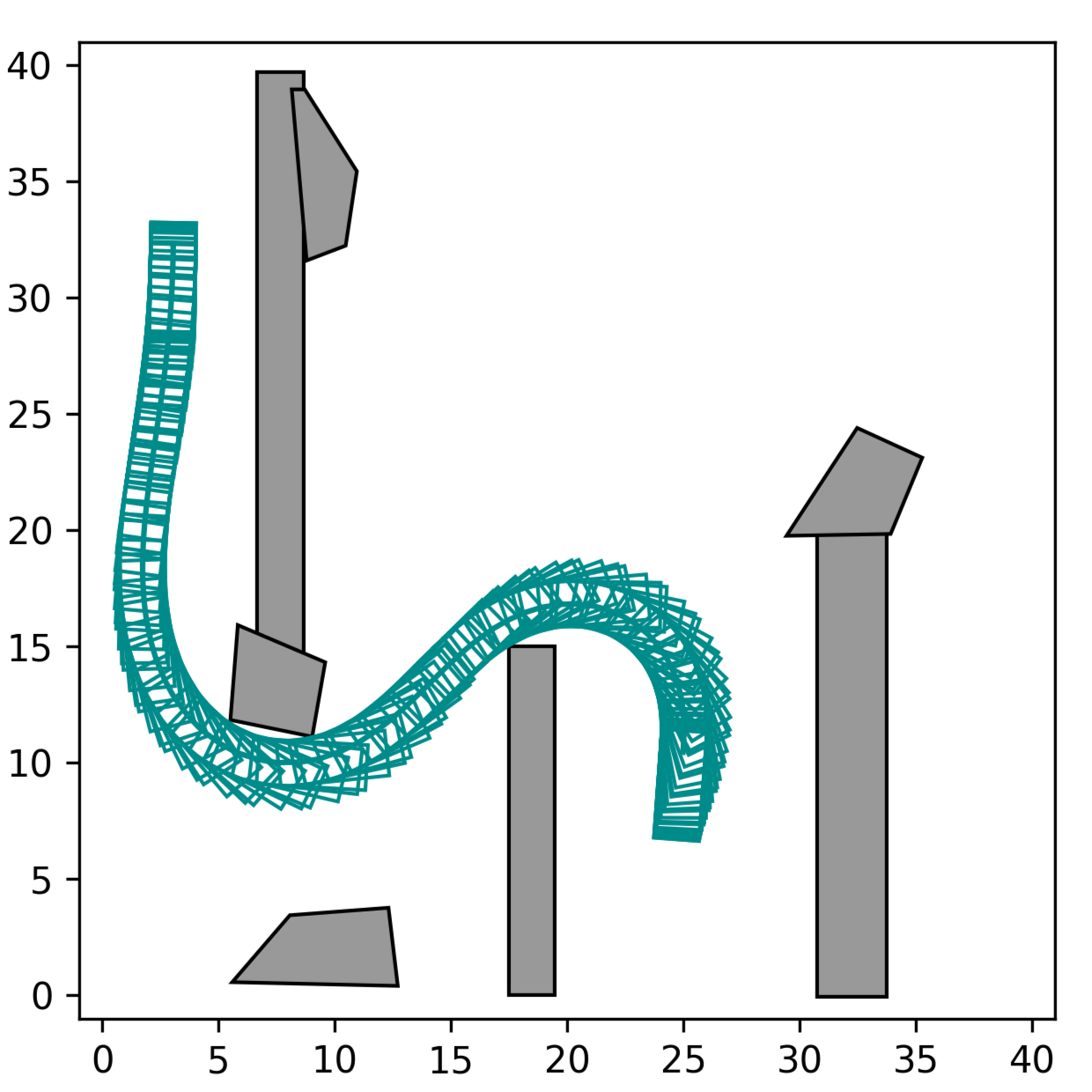}}
\hfil
\subfloat[]
{\includegraphics[width=4.0cm]{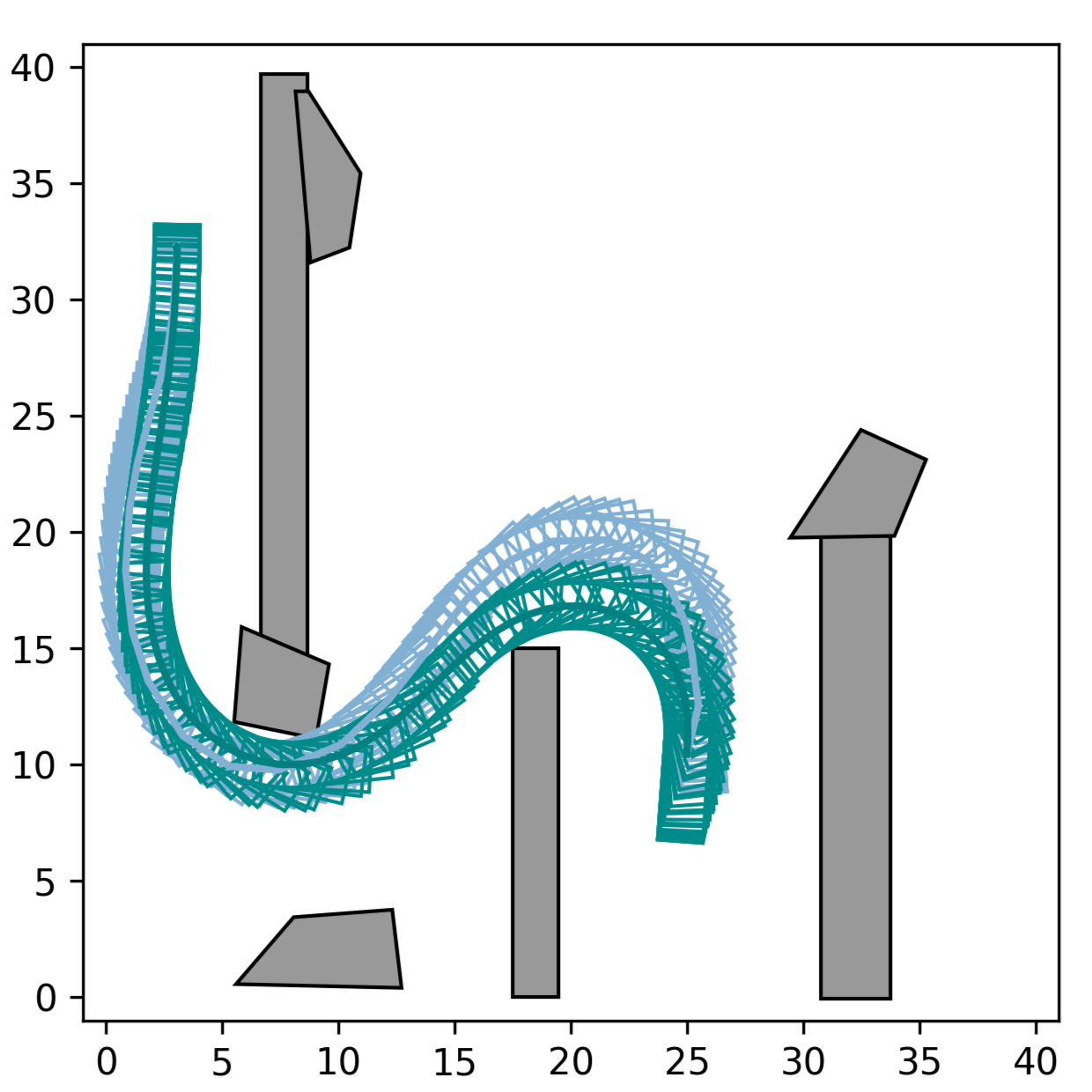}}
\hfil
\caption{Trajectory generation and optimization process}
\label{fig:net_pre}
\end{figure}

Due to space limitations, we use \ref{fig:test_map}(a) as a representative example to illustrate the trajectory generation process in detail, as shown in Fig. \ref{fig:net_pre}. Given the known environmental information, we use the trained neural network model to predict the vehicle trajectory. The inputs include environmental information, starting point information, and target point information. Through the GNN, the model outputs predicted trajectory points. In Fig. \ref{fig:net_pre}, Fig. \ref{fig:net_pre}(a) represents the predicted path points, consisting of 30 discrete points. Each discrete point includes values for the vehicle's orientation angle $\theta$, steering angle $\varphi$, and velocity $v$. The steering angle velocity $w$ and acceleration $a$ required for the OCP problem are calculated from the steering angle and velocity. To match the number of predicted points with the optimization points required by the OCP, we perform linear interpolation between every two trajectory points to add three additional trajectory points, as shown in \ref{fig:net_pre}(b). The interpolated trajectory in \ref{fig:net_pre}(b) includes $x$, $y$, $\theta$, $\varphi$, $v$, $w$ and $a$, which will serve as the initial solution for the IPOPT iterative optimization. \ref{fig:net_pre}(c) shows the results after solving the OCP, and \ref{fig:net_pre}(d) compares the paths before and after optimization.

\begin{figure}[htbp]
\centering
\subfloat[]
{\includegraphics[width=3.1cm]{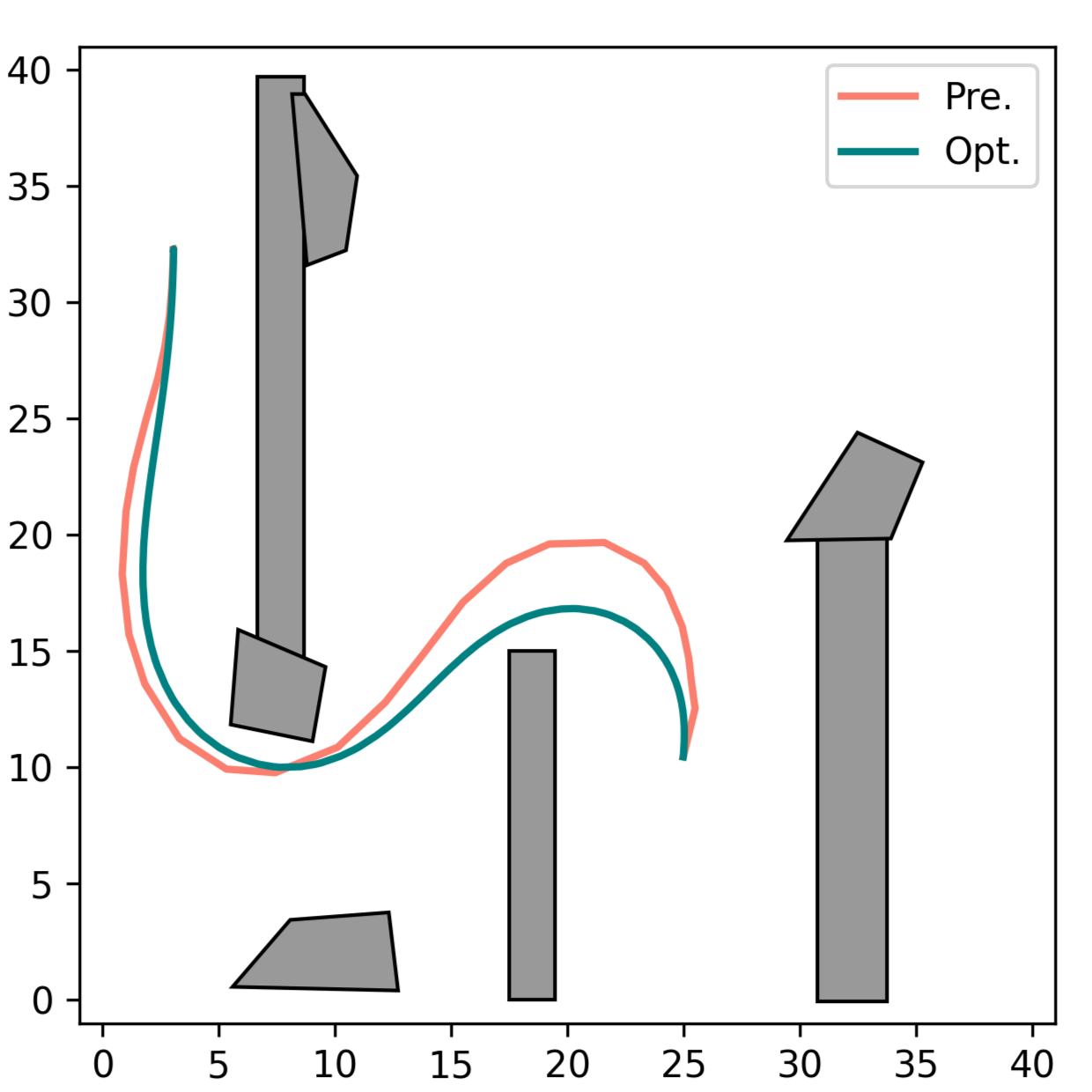}}
\hfil
\subfloat[]
{\includegraphics[width=4.0cm]{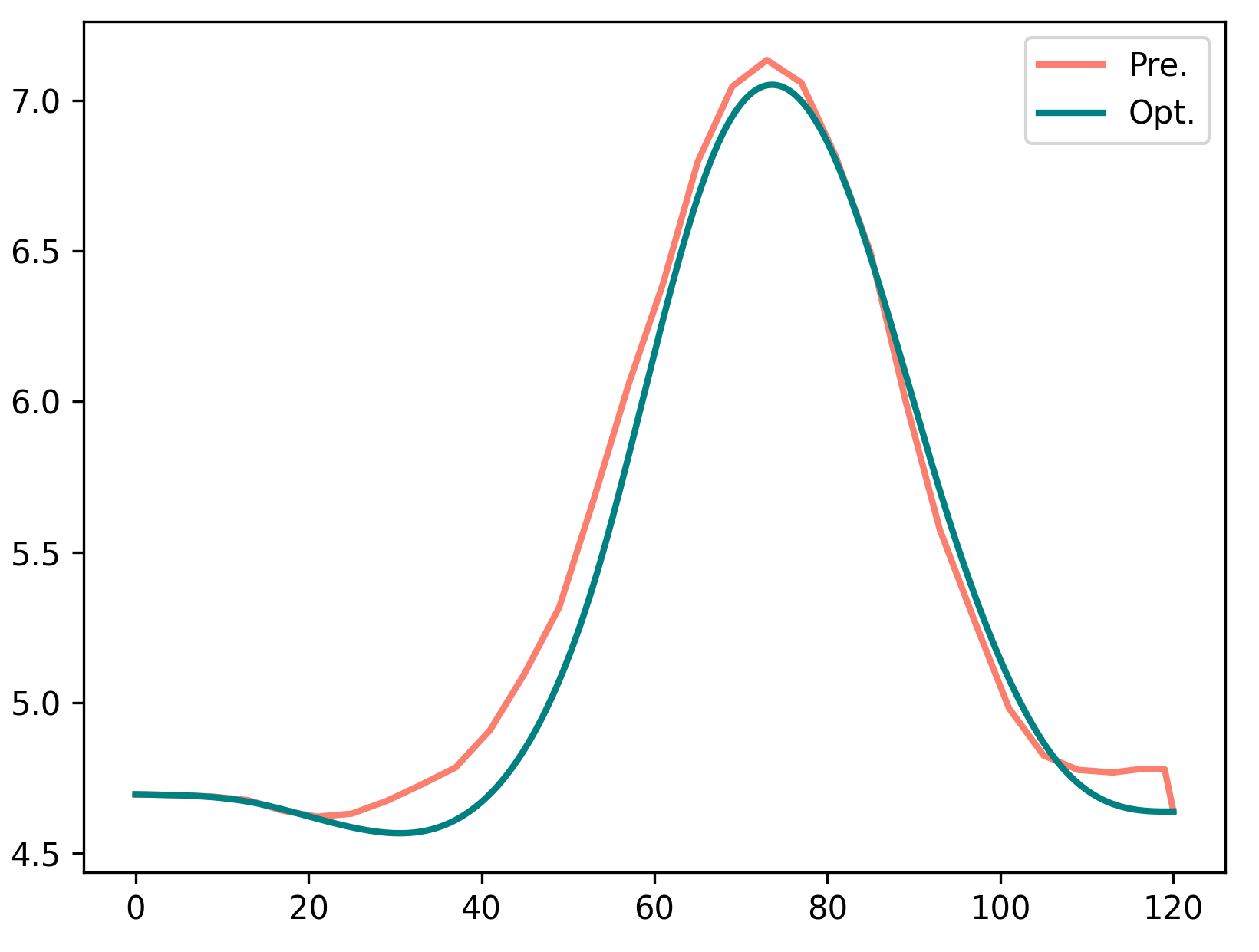}}
\hfil
\subfloat[]
{\includegraphics[width=4.0cm]{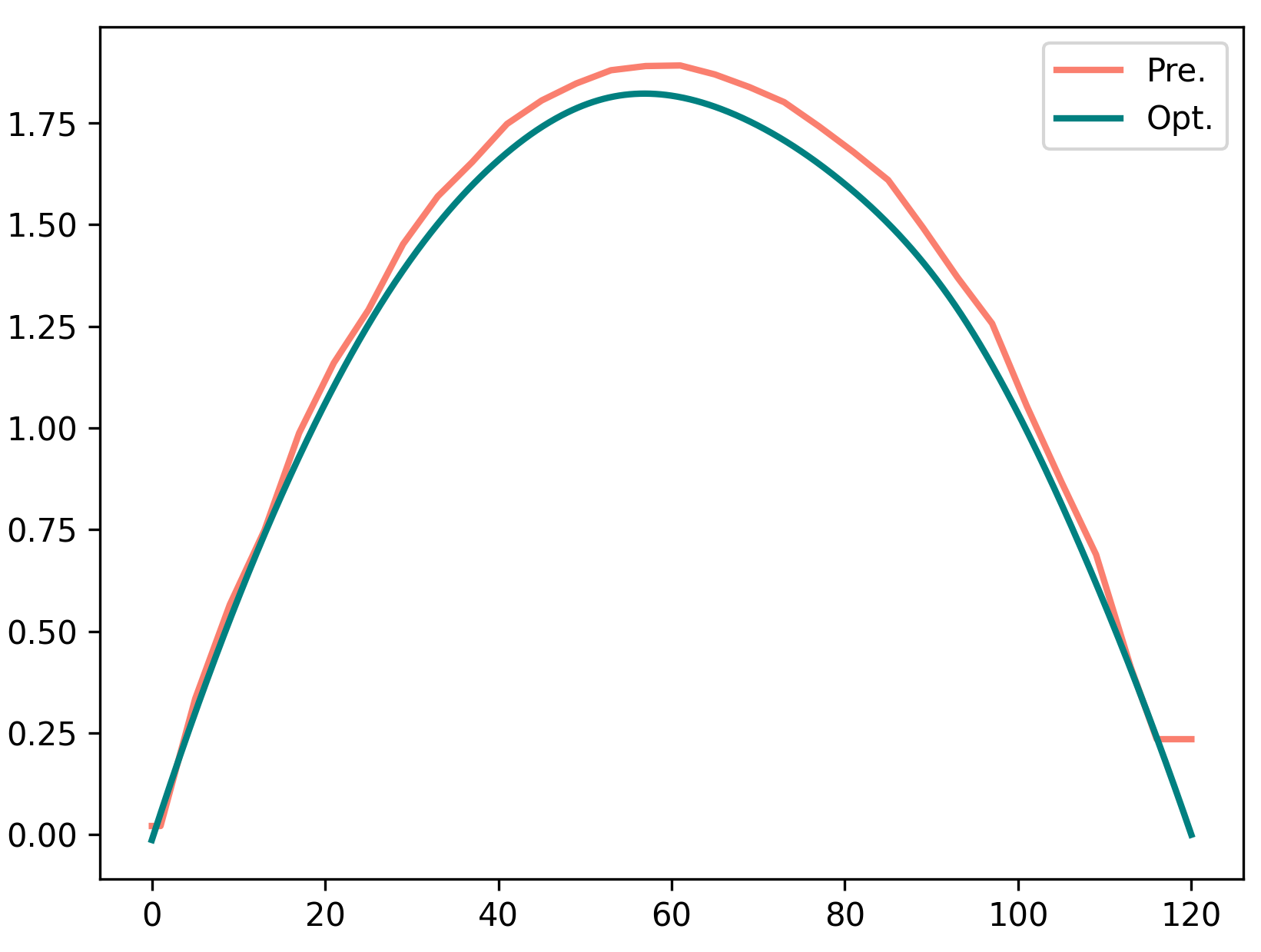}}
\hfil
\subfloat[]
{\includegraphics[width=4.0cm]{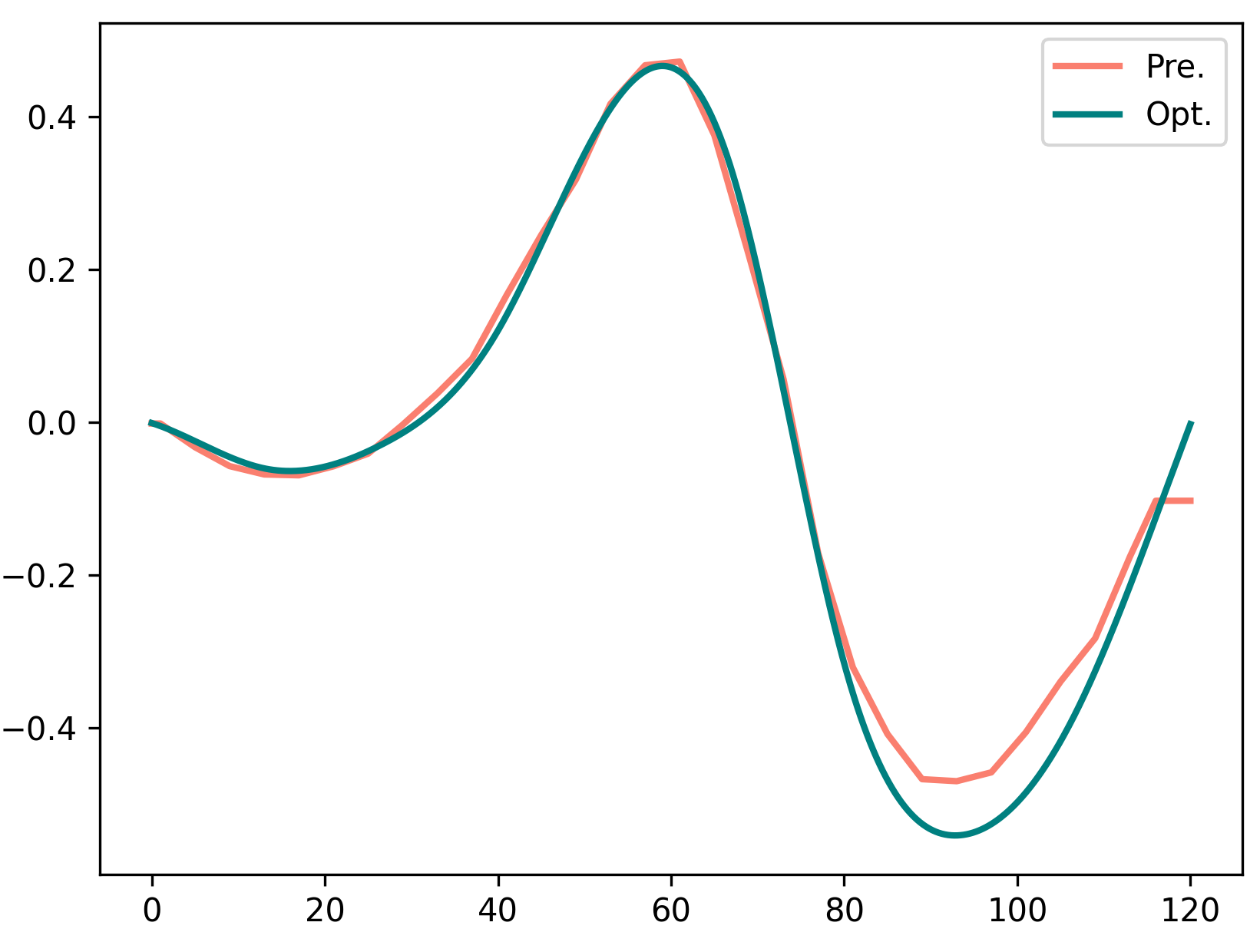}}
\hfil
\caption{Comparison between GNN trajectory prediction and optimization trajectory}
\label{fig:detail_comparison}
\end{figure}

Additionally, Fig.\ref{fig:detail_comparison} provides a detailed comparison of the trajectory parameters between the predicted trajectory and the local optimal trajectory from Fig.\ref{fig:net_pre}(d). Fig.\ref{fig:detail_comparison}(a), Fig.\ref{fig:detail_comparison}(b), Fig.\ref{fig:detail_comparison}(c), and Fig.\ref{fig:detail_comparison}(d) respectively show the comparisons of the path, heading angle, velocity, and steering angle between the predicted trajectory and the local optimal trajectory.

It can be observed that the prediction from the neural network closely matches the results after OCP optimization. Therefore, the network output, after interpolation processing, can serve as the initial solution for the OCP problem.

\subsection{Comparison with mainstream planners}
In this section, the neural network combined with optimal control proposed in this paper is compared with current mainstream non-structured road trajectory planning algorithms combined with optimal control, including Hybrid A* and RRT* combined with optimal control, in both the initial trajectory generation and trajectory optimization stages. The testing environment is shown in Figure 6. The specific comparison results are shown in Table \ref{table:Comparison}.
\begin{table}[!htbp]
\caption{Performance Comparison}
\label{table:Comparison}
\begin{center}
\begin{tabular}{ 
>{\centering\arraybackslash}m{0.8cm} 
>{\centering\arraybackslash}m{2.0cm} 
>{\centering\arraybackslash}m{1.2cm} 
>{\centering\arraybackslash}m{0.8cm} 
>{\centering\arraybackslash}m{0.8cm}}
 \hline
 Cases &Method &Planing time(s)&Opt. time(s) & Total time(s)\\ 
 \hline
 \multirow{3}*{map(a)} & Hybrid A*-OCP & 4.487 & \textbf{5.549}& 10.036\\ 
  & RRT*-OCP &\textbf{0.167}  & 23.641& 23.808 \\ 
  & GNN-OCP & 0.695 & 6.652& \textbf{8.347}\\ 
 \multirow{3}*{map(b)} & Hybrid A*-OCP & 3.988 & \textbf{4.870}& 8.858\\ 
  & RRT*-OCP & \textbf{0.284} & 15.904& 16.188\\ 
  & GNN-OCP & 0.709 & 6.956& \textbf{7.665}\\ 
 \multirow{3}*{map(c)} & Hybrid A*-OCP & 1.089 & 8.134& 9.223\\ 
  & RRT*-OCP & \textbf{0.162} & 9.675& 9.837\\ 
  & GNN-OCP& 0.691 & \textbf{2.874}& \textbf{3.565}\\  
 \multirow{3}*{map(d)} & Hybrid A*-OCP & 1.756 & \textbf{8.137}& 9.893\\ 
  & RRT*-OCP & \textbf{0.152} & 13.948& 14.100\\ 
  & GNN-OCP & 0.687 & 8.507& \textbf{9.194}\\ 
\hline
\end{tabular}
\end{center}
\end{table}

In the first stage of planning, the initial solution is provided by the algorithm or the neural network. It can be observed that the RRT* algorithm exhibits the highest efficiency and the shortest time consumption due to its disregard for the vehicle's kinematic constraints. The Hybrid A* algorithm consumes more time, particularly in environments like map (a) and map (b) where the starting point is relatively distant from the target, as graph search algorithms require step-by-step expansion from the starting point to the endpoint, resulting in lower efficiency with longer distances. In comparison, the time consumption of the GNN remains stable at around 700ms. This stability arises from the fact that, given the same computational unit, the inference speed of the neural network is only determined by the network structure.

In the second stage of planning, i.e., the OCP optimization stage, the Hybrid A*-OCP exhibits the shortest time consumption among the map(a), map(b), and map(d) experimental scenarios, while the RRT*-OCP has the longest time consumption among all experimental scenarios. The GNN-OCP slightly surpasses the Hybrid A*-OCP. In this stage, the duration of OCP optimization largely depends on the quality of the initial solution, the closer the initial solution to the optimal solution, the lower the time consumption and the faster the optimization rate. Among the four proposed experimental scenarios, the GNN-OCP proposed in this paper exhibits the shortest total time, significantly improving trajectory planning efficiency.

\begin{figure}[htbp]
\centering
\subfloat[Position]
{\includegraphics[width=4.0cm]{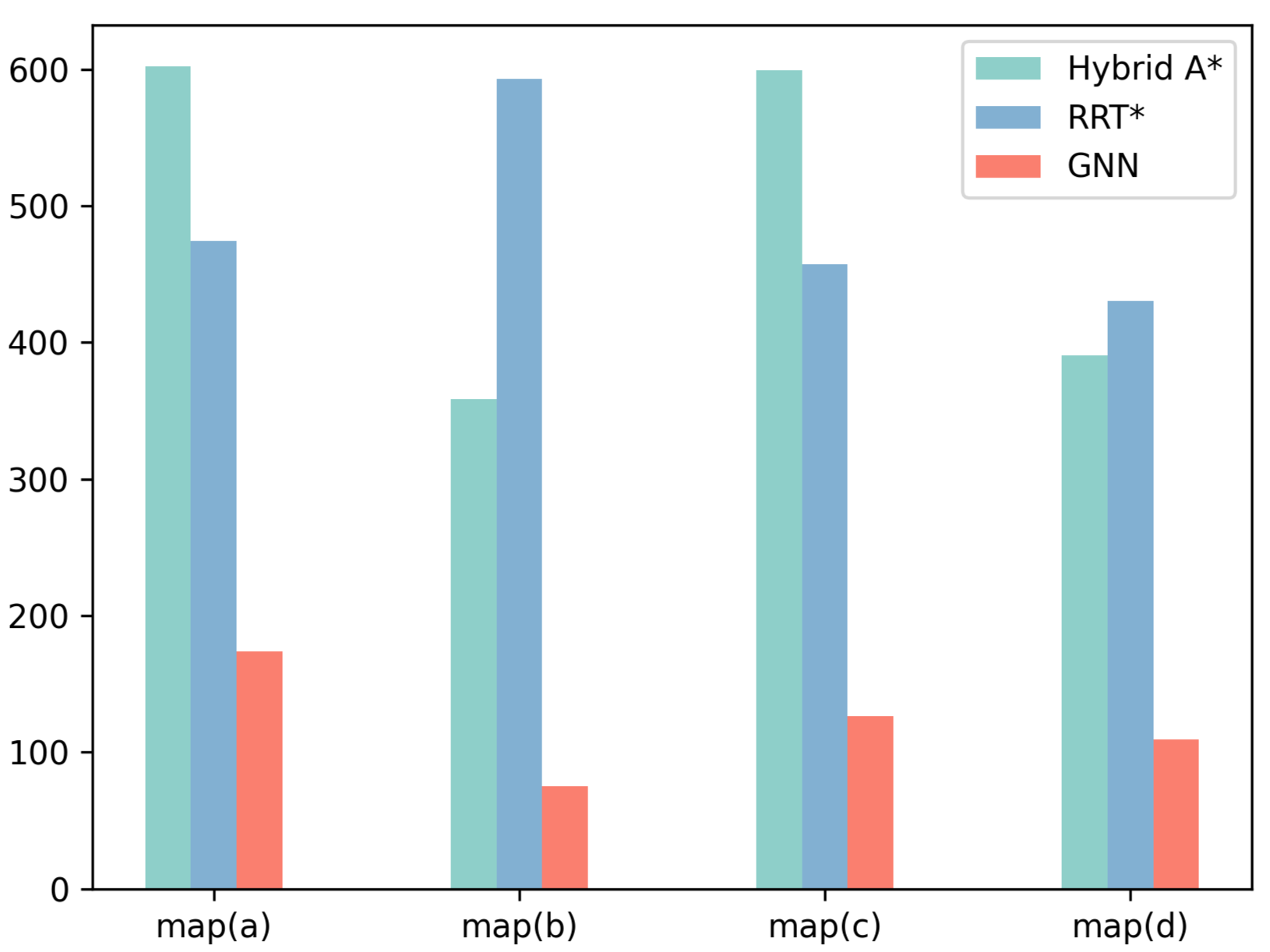}}
\hfil
\subfloat[Heading angle]
{\includegraphics[width=4.0cm]{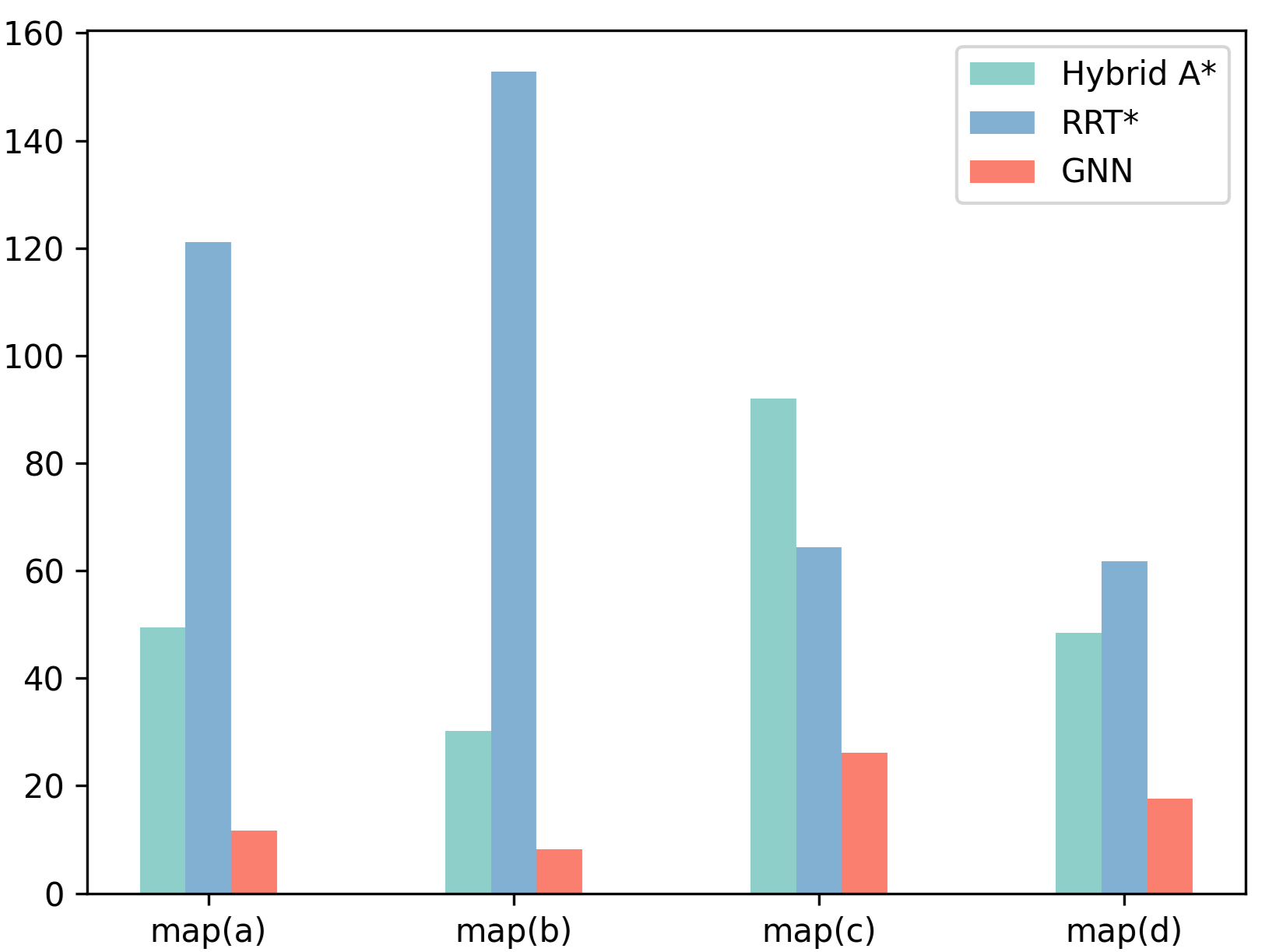}}
\hfil
\subfloat[Speed]
{\includegraphics[width=4.0cm]{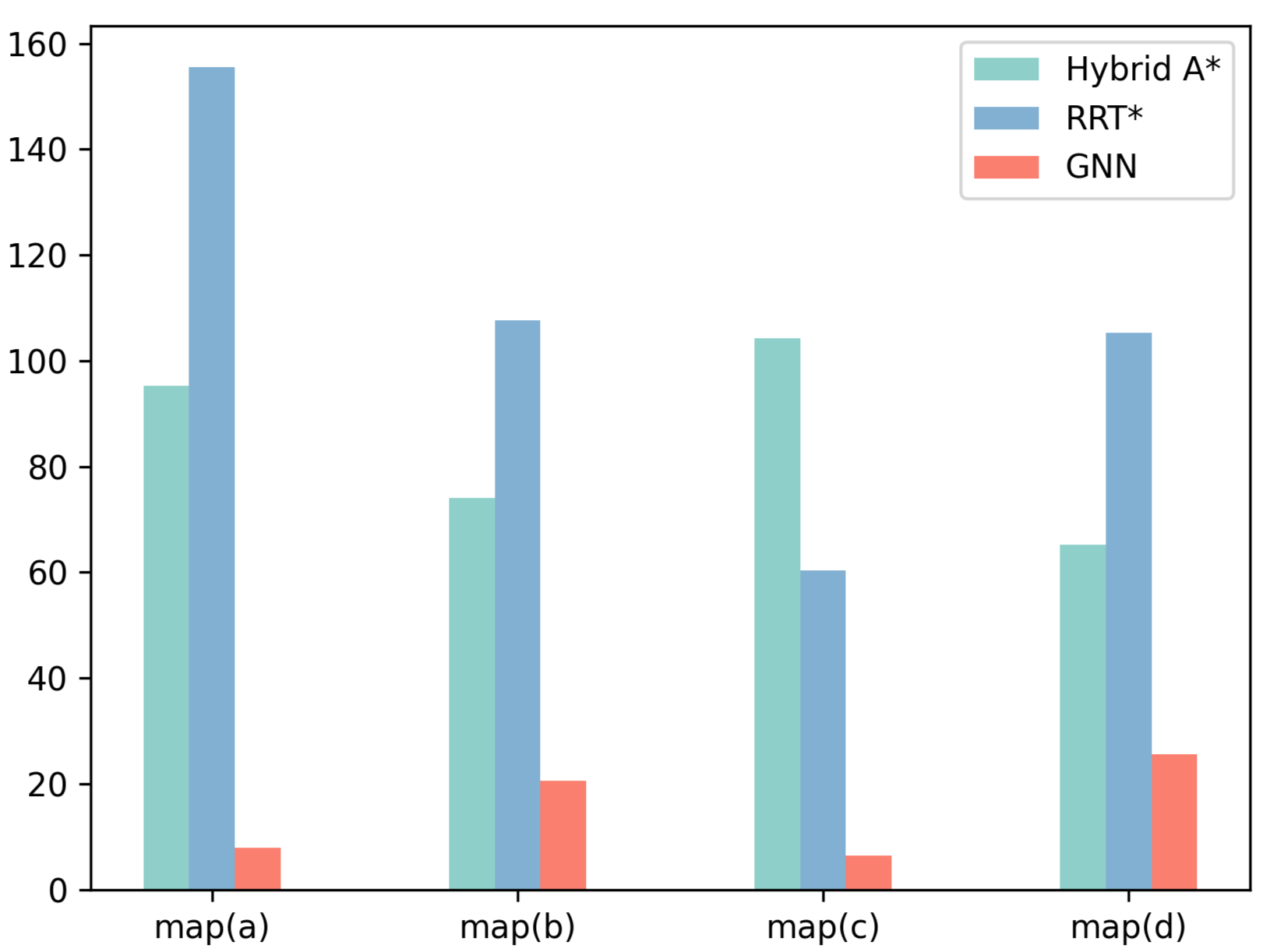}}
\hfil
\subfloat[Steering angle]
{\includegraphics[width=4.0cm]{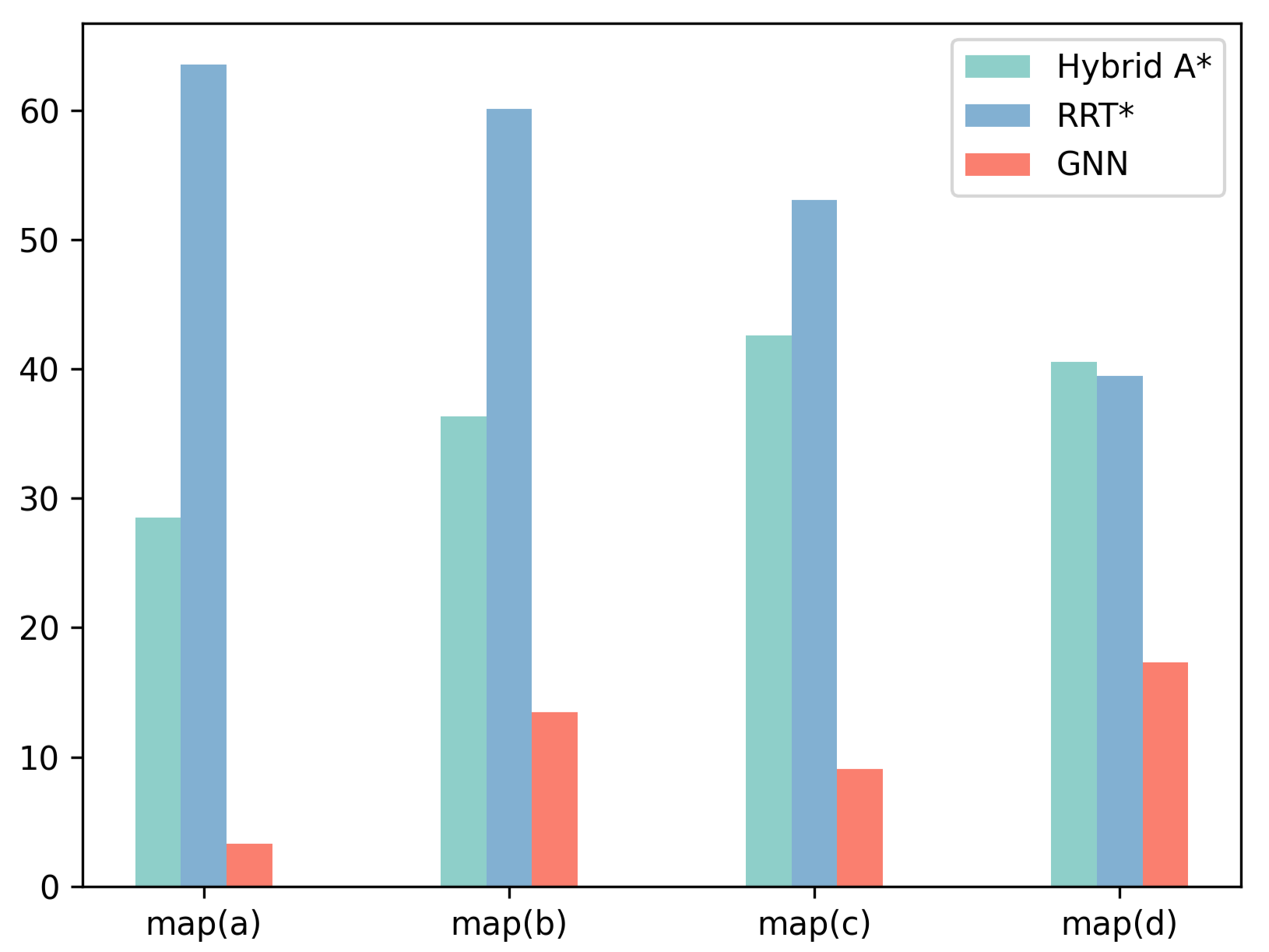}}
\hfil
\caption{The difference distance between the first stage planning trajectory and the local optimal solution}
\label{fig:disparity}
\end{figure}

To delve deeper into the performance analysis of the algorithms presented in Table \ref{table:Comparison}, we computed the differences between the planned trajectories in the first stage and the local optimal solutions in the second stage for each algorithm, as shown in Fig. \ref{fig:disparity}. Fig. \ref{fig:disparity}(a) illustrates the cumulative sum of distances between the planned trajectory and the local optimal trajectory for each trajectory point. Figures \ref{fig:disparity}(b), \ref{fig:disparity}(c), and \ref{fig:disparity}(d) depict the cumulative sums of differences for each trajectory point in terms of $\theta$, $v$, and $\varphi$, respectively. It can be seen that the trajectory points predicted by the GNN are significantly closer to the locally optimal solution compared to the other algorithms. However, as shown in Table \ref{table:Comparison}, in the experimental environments of map(a), map(b), and map(d), the optimization time of GNN-OCP is slightly higher than that of Hybrid A*. This is because the GNN faces a similar issue as the RRT* algorithm: it cannot completely satisfy the vehicle's kinematic constraints during trajectory prediction. The GNN predicts $x$, $y$, $\theta$, $\varphi$, and $v$ independently, ensuring each value adheres to its maximum and minimum constraints but without establishing correlations to meet the vehicle's kinematic equations. Compared to RRT*, the GNN's predicted trajectories are closer to the feasible solution range dictated by the vehicle's kinematic constraints, resulting in performance that is nearly on par with Hybrid A*-OCP.

\subsection{Parameter analysis and model defects}
As mentioned in Section \ref{Proposed Trajectory Planner}, we employed a simple yet effective network. However, there is room for improvement in the network structure, and the algorithm parameters can be adjusted in various ways. This section will provide a brief analysis of the algorithm parameters.

\begin{figure}[htbp]
\centering
\subfloat[]
{\includegraphics[width=6.5cm]{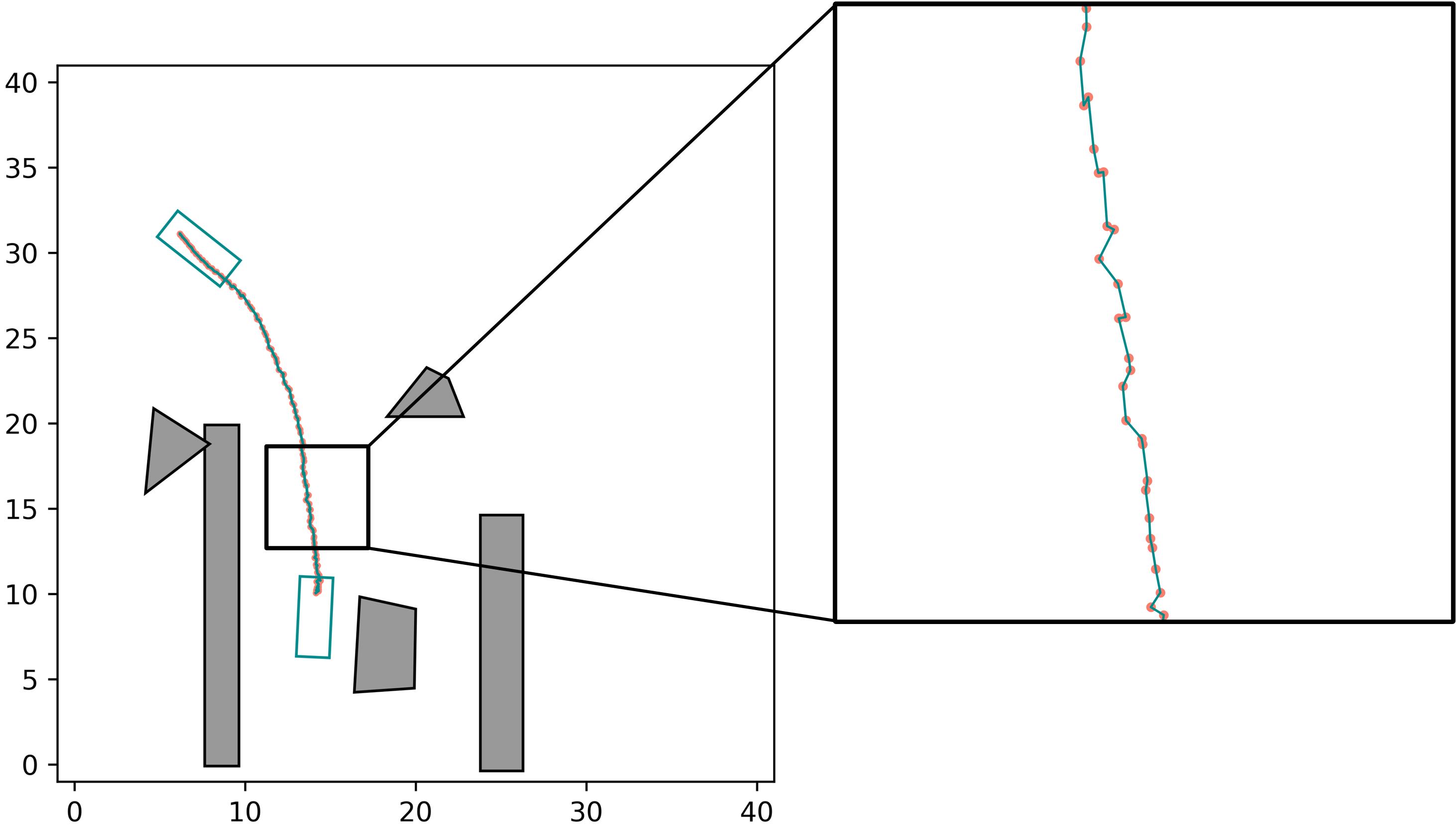}}
\hfil
\subfloat[]
{\includegraphics[width=6.5cm]{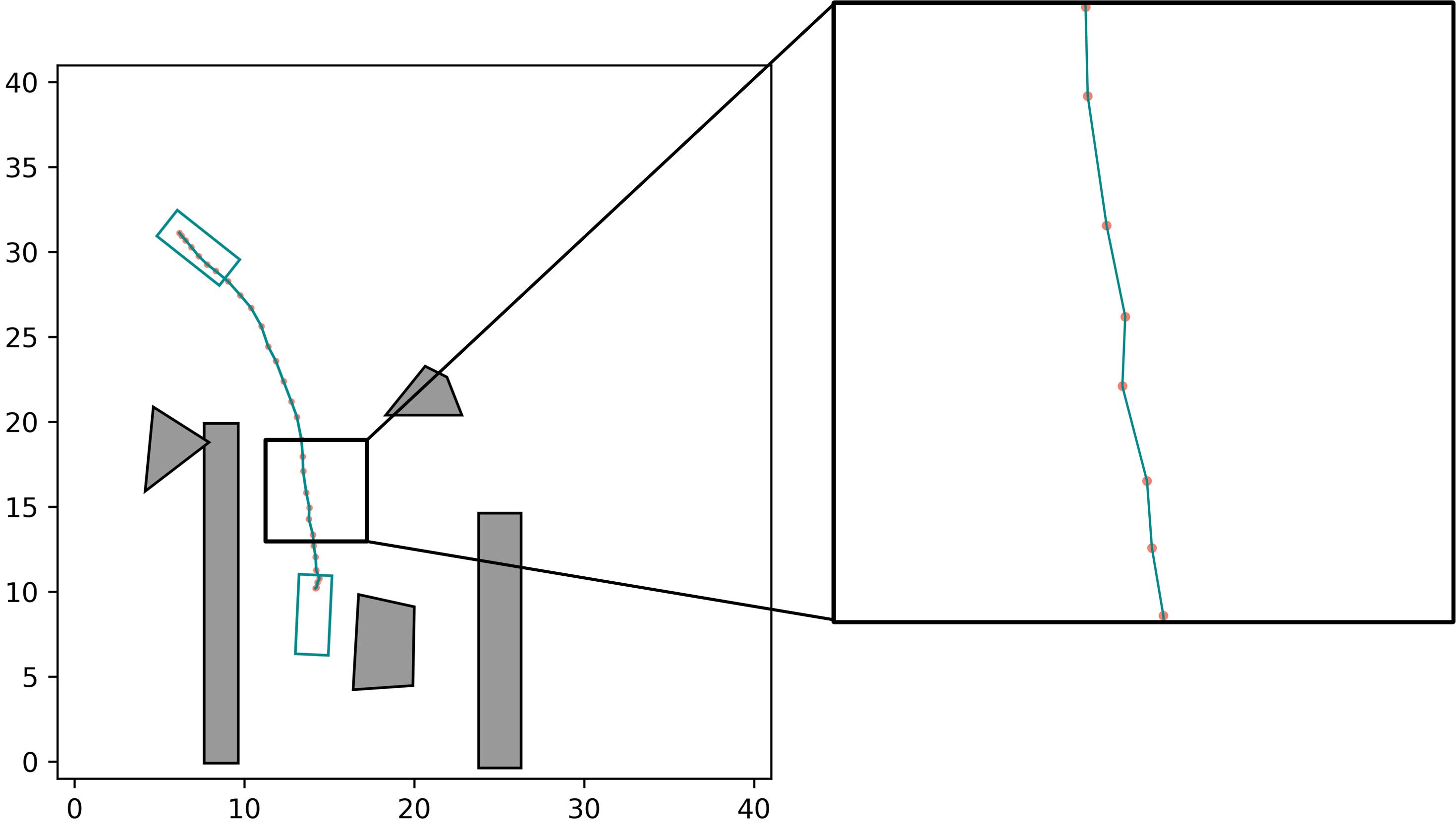}}
\hfil
\subfloat[]
{\includegraphics[width=6.5cm]{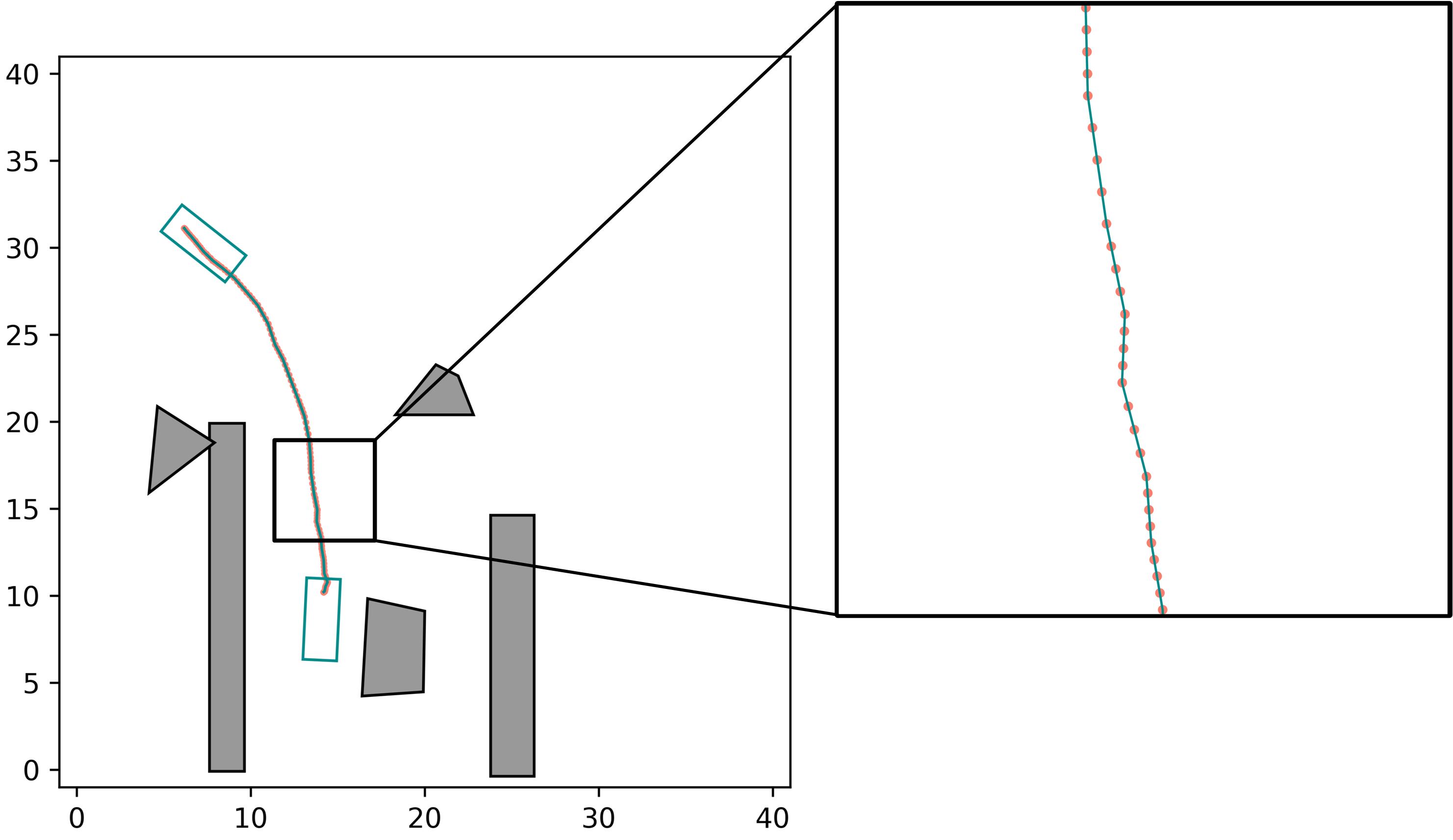}}
\hfil
\caption{Prediction node interpolation processing}
\label{fig:pre_out}
\end{figure}

In selecting the number of trajectory points predicted by the neural network, we chose to use 30 prediction points. Due to inherent prediction errors in the neural network, the number of prediction points determines the smoothness of the predicted trajectory and the optimization rate. As shown in Fig. \ref{fig:pre_out}, Fig. \ref{fig:pre_out}(a) illustrates a trajectory with 120 predicted points, where the trajectory's smoothness is compromised due to prediction errors. When the number of prediction points is reduced, as shown in Fig. \ref{fig:pre_out}(b), and linear interpolation is applied between the prediction points, the final result is displayed in Fig. \ref{fig:pre_out}(c).


\begin{figure}[htbp]
\centering
\subfloat[]
{\includegraphics[width=3.5cm]{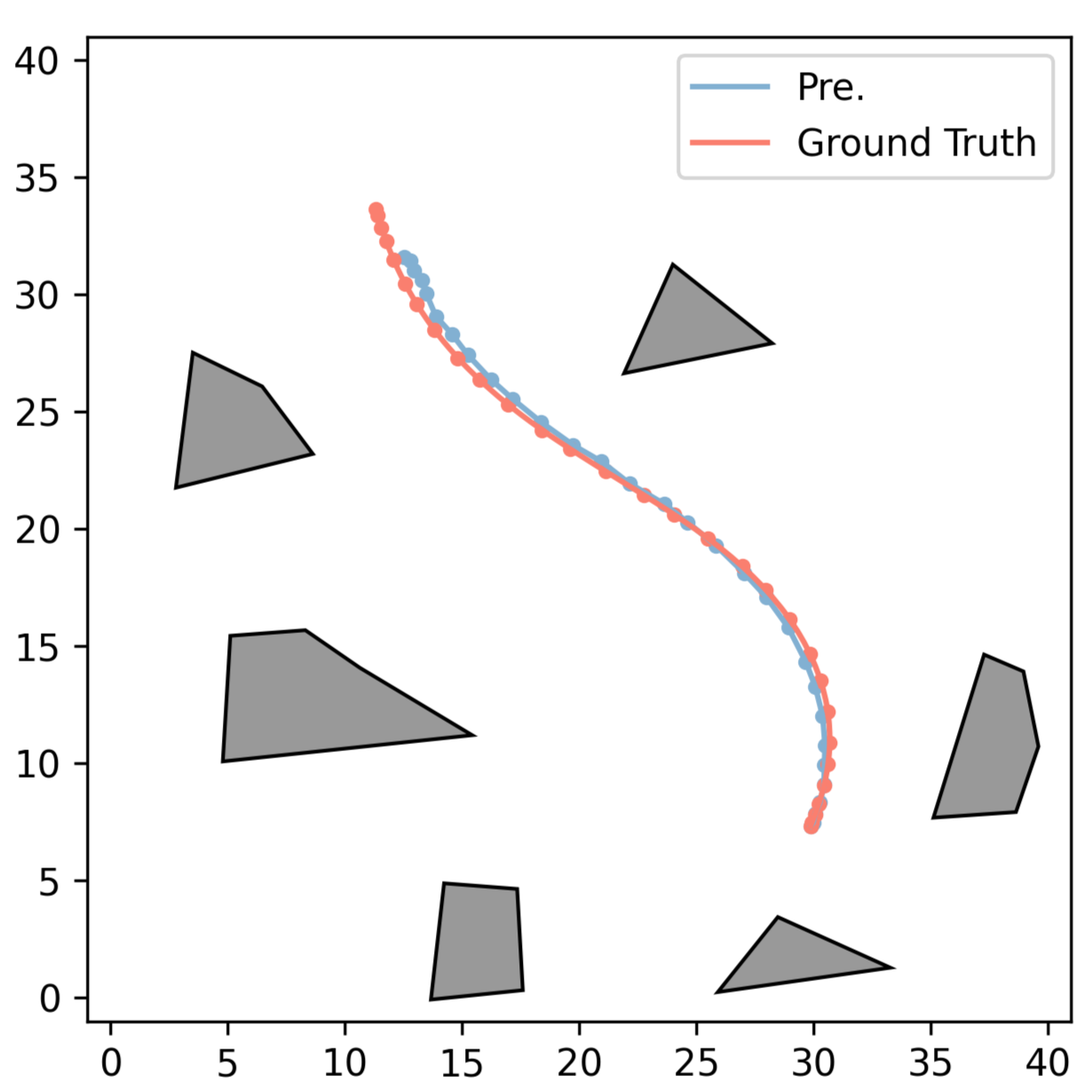}}
\hfil
\subfloat[]
{\includegraphics[width=4.5cm]{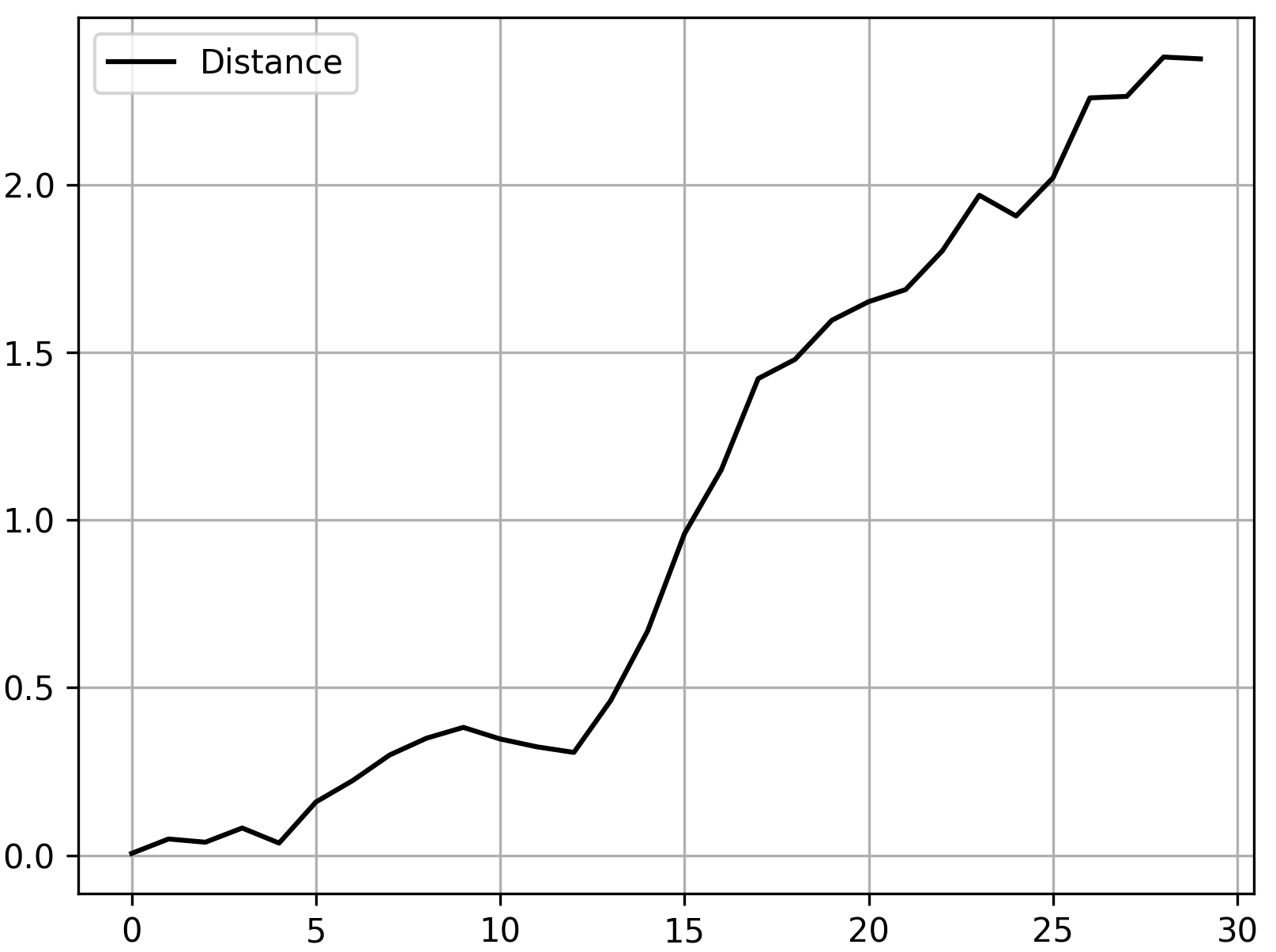}}
\hfil
\caption{Change in distance between predicted trajectory and actual trajectory}
\label{fig:pre_ground_disparity}
\end{figure}

When using neural networks, their performance tends to deteriorate when handling tasks significantly different from the training dataset, which is an inherent drawback of neural networks. In this study, the prediction error for trajectories increases with the length of the trajectory; the further from the starting point, the greater the error. As shown in Fig. \ref{fig:pre_ground_disparity},  Fig. \ref{fig:pre_ground_disparity}(a) illustrates the predicted trajectory and the positions of the trajectory points.  Fig. \ref{fig:pre_ground_disparity}(b) depicts the distance between each predicted trajectory point and its corresponding true trajectory label point, starting from the initial point. The distance gradually increases with the length of the trajectory. Nevertheless, as long as the error remains within a reasonable range, it will not affect the calculation of the locally optimal trajectory during the optimization phase.

\section{Conclusion}
\label{Conclusion}
This paper proposes a two-stage trajectory planning method based on GNN and numerical optimal control. In the first stage, the GNN efficiently extracts environmental task information to predict an initial trajectory. In the second stage, optimal control computation is employed to optimize the trajectory, ensuring it meets vehicle kinematic constraints and collision avoidance requirements. This method simplifies the traditional trajectory planning process. Unlike conventional planning algorithms that require separate steps for path planning, speed planning, and trajectory optimization, the GNN-based initial planning benefits from the simplicity and efficiency of end-to-end trajectory planning. This reduces the need for extensive parameter tuning in traditional planning algorithms while enhancing the overall planning efficiency.

The neural network structure proposed in this paper is simple and efficient, but there is significant room for improvement. Future work will consider larger datasets and deeper network structures to enhance the predictive capabilities of the neural network. In terms of prediction results, this paper uses parallel multilayer perceptrons to predict different trajectory parameters. While this approach achieves good results, it does not simultaneously incorporate different results into the vehicle kinematic model, which reduces the optimization process speed. Nonetheless, in the initial trajectory acquisition stage, the algorithm demonstrates sufficient efficiency to reduce the overall running time and simplify the workflow. The planning framework provided in this paper also offers a novel perspective for end-to-end trajectory planning: using numerical optimization methods to improve the trajectory quality predicted by neural networks.

\bibliographystyle{IEEEtran}
\bibliography{arxiv}
\newpage
\vfill

\end{document}